\pdfoutput=1

\documentclass[11pt]{article}

\usepackage[final]{acl}

\usepackage{times}
\usepackage{latexsym}
\usepackage{amsmath}
\usepackage{cleveref}

\usepackage[T1]{fontenc}

\usepackage[utf8]{inputenc}

\usepackage{microtype}

\usepackage{inconsolata}

\usepackage{paralist}
\usepackage[normalem]{ulem}
\usepackage{color}
\usepackage{booktabs}

\usepackage{xcolor}
\usepackage{colortbl}
\definecolor{lightBlue}{HTML}{A7C8DE}
\definecolor{DarkBlue}{HTML}{5790C0}
\definecolor{DarkGreen}{HTML}{3d6f28}
\definecolor{DarkRed}{HTML}{b02102}
\definecolor{HighlightBlue}{RGB}{220,230,241}
\definecolor{HighlightRed}{HTML}{B22222}

\usepackage{graphicx}

\usepackage{array}  
\usepackage{multirow} 
\usepackage{balance}
\usepackage{float}

\usepackage[breakable]{tcolorbox}

\title{Unmasking Deceptive Visuals: Benchmarking Multimodal Large Language Models on Misleading Chart Question Answering}

\author{
 \textbf{Zixin Chen\textsuperscript{1}},
 \textbf{Sicheng Song\textsuperscript{1}$^{\heartsuit}$},
 \textbf{Kashun Shum\textsuperscript{1}},
 \textbf{Yanna Lin\textsuperscript{1}},
\\
 \textbf{Rui Sheng\textsuperscript{1}},
 \textbf{Weiqi Wang\textsuperscript{1}},
 \textbf{Huamin Qu\textsuperscript{1}}
\\
 \textsuperscript{1}The Hong Kong University of Science and Technology
\\
\texttt{\{zchendf,ksshumab,ylindg,rshengac,wwangbw\}@connect.ust.hk} \\
\texttt{csescsong@ust.hk,huamin@cse.ust.hk}
}

\begin{document}

\definecolor{titleblockcolor}{HTML}{353535}
\definecolor{textblockcolor}{HTML}{FFFFFF}
\newenvironment{block}[2][]{
  \begin{tcolorbox}[adjusted title=#2, fonttitle={\normalsize\bfseries}, colback={textblockcolor}, colframe={titleblockcolor}, coltitle={white}, arc=0pt,
  outer arc=0pt, left=1pt, right=1pt, fontupper=\normalsize, #1, breakable]
}{\end{tcolorbox}}
\newtcbox{\highlight}[1][red]
  {on line, arc = 0pt, outer arc = 0pt,
    colback = #1!10!white, colframe = #1!50!black,
    boxsep = 0pt, left = 1pt, right = 1pt, top = 2pt, bottom = 2pt,
    boxrule = 0pt, bottomrule = 1pt, toprule = 1pt}

\maketitle

\def\thefootnote{$^{\heartsuit}$}\footnotetext{The corresponding author.}
\def\thefootnote{\arabic{footnote}}

\begin{abstract}
Misleading visualizations, which manipulate chart representations to support specific claims, can distort perception and lead to incorrect conclusions. Despite decades of research, they remain a widespread issue, posing risks to public understanding and raising safety concerns for AI systems involved in data-driven communication. While recent multimodal large language models (MLLMs) show strong chart comprehension abilities, their capacity to detect and interpret misleading charts remains unexplored. We introduce Misleading ChartQA benchmark, a large-scale multimodal dataset designed to evaluate MLLMs on misleading chart reasoning. It contains 3,026 curated examples spanning 21 misleader types and 10 chart types, each with standardized chart code, CSV data, multiple-choice questions, and labeled explanations, validated through iterative MLLM checks and expert human review. We benchmark 24 state-of-the-art MLLMs, analyze their performance across misleader types and chart formats, and propose a novel region-aware reasoning pipeline that enhances model accuracy. Our work lays the foundation for developing MLLMs that are robust, trustworthy, and aligned with the demands of responsible visual communication.
\end{abstract}

\section{Introduction}

\begin{figure}[!ht]
    \centering
    \includegraphics[width=\linewidth]{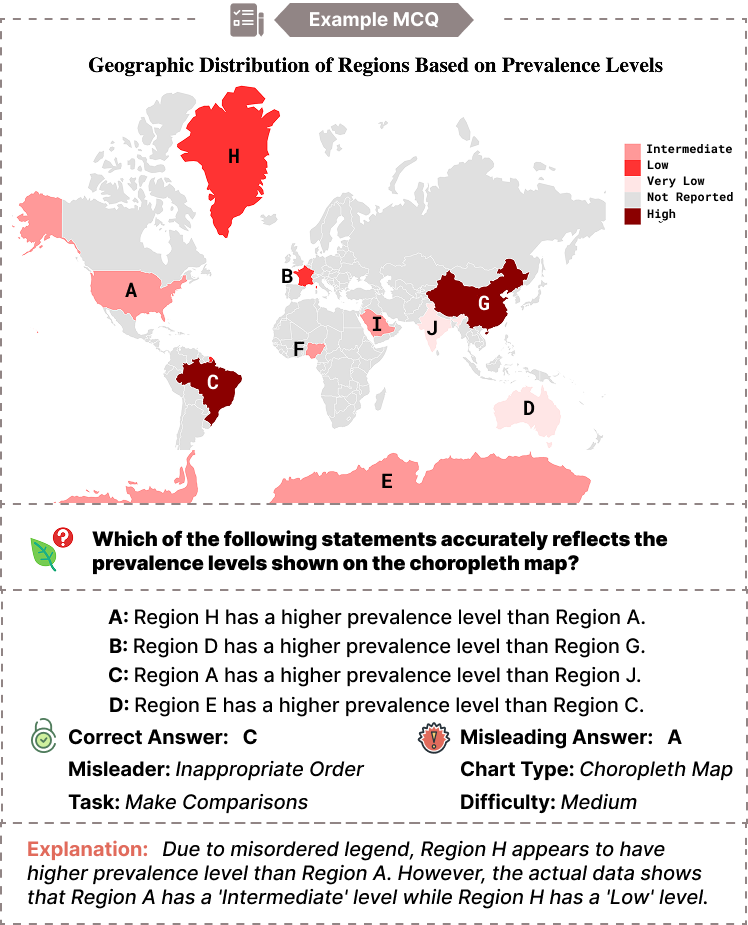}
    \caption{An example multiple-choice question (MCQ) from our benchmark. Each MCQ includes a misleading chart, a question, multiple answer options, the correct answer and a set of labels. A detailed explanation is also provided to illustrate the chart’s misleading aspects.}
    \label{fig:example}
\end{figure}

Misleading visualizations have long posed challenges in chart comprehension and public communication~\cite{tufte1983visual}. As early as the 1950s, the influential book \textit{How to Lie with Statistics} illustrated how selectively constructed charts could distort data and manipulate public perception~\cite{huff1954lie}. Despite decades of awareness, misleading designs remain common today. For example, in 2020, the Georgia Department of Public Health released a COVID-19 bar chart sorted by case count rather than date, falsely implying a decline in infections~\cite{mcfall-johnsen-2020} (\cref{fig:misleading_samples} A). Another widely recognized example is the standard world map under Mercator Projection (\cref{fig:misleading_samples} B), which distorts country sizes by exaggerating areas near the poles~\cite{kennedy2000understanding, obrien-2024}. These real-world cases illustrate how charts can subtly mislead audiences, posing risks to public understanding and highlighting the importance of trustworthy data communication.

Recent advances in multimodal large language models (MLLMs) have shown strong performance on chart-related tasks such as question answering~\cite{xia2024chartx, masry2022chartqa}, captioning~\cite{huang2023lvlms, rahman2023chartsumm}, and structure extraction~\cite{chen2024onechart}. However, most existing work focuses on factual interpretation and overlooks the critical challenge of detecting and reasoning about misleading visual content. Although this issue has long been recognized in the visualization literature~\cite{tufte1983visual, ge2023calvi}, it remains largely unaddressed in the context of MLLMs. 

As MLLMs are increasingly applied in high-stakes domains such as news summarization, policy analysis, and scientific communication, the ability to recognize and resist visual manipulation becomes essential. Such robustness is crucial not only for combating misinformation but also for ensuring responsible AI deployment aligned with user intent, legal norms, and societal values. Despite its importance, progress on this problem has been limited, which we attribute to three main challenges: (1) the theoretical difficulty of defining and organizing diverse misleading features and aligning them with corresponding chart formats; (2) the complexity and cognitive effort required to design high-quality question–answer pairs that capture realistic misleading scenarios; and (3) the expert labor needed for accurate annotation and validation.


To address this gap, we present the Misleading ChartQA benchmark, a large-scale multimodal dataset for evaluating MLLMs' ability to identify and reason about misleading charts. Our work builds on theoretical foundations that define common misleading features (misleaders)~\cite{borner2019data, lo2022misinformed, lan2024came} and multiple-choice question (MCQ) frameworks used to assess human interpretation~\cite{lee2016vlat, cui2023adaptive, ge2023calvi}.


We collaborated with data visualization experts to develop a comprehensive misleader taxonomy (\cref{fig:misleaderTaxonomy}), covering 60 unique (misleader, chart type) pairs across 21 misleaders and 10 chart types (\cref{fig:misleaderDefinition}). For each pair, experts authored 2–3 well-defined examples, resulting in a total of 155 seed MCQs, which were standardized into D3.js~\cite{bostock2011d3} visualizations, CSV data, and labeled JSON formats. Using automated expansion and expert review by 20 trained reviewers, we constructed a dataset of 3,026 curated misleading chart MCQs. We benchmark 24 state-of-the-art MLLMs and systematically analyze their performance across misleader types, chart formats, and error patterns, based on the testing set. To support future progress, we propose a Region-Aware Misleader Reasoning pipeline that enhances MLLM performance by explicitly guiding attention to misleading chart regions.

\section{Misleading ChartQA Benchmark}

In this section, we describe the construction of the Misleading ChartQA dataset, which involves four main stages: (1) Misleader Taxonomy Construction, (2) Seed MCQ Design, (3) MCQ Augmentation and Iterative Refinement, and (4) Intensive Expert Validation.

\subsection{Misleader Taxonomy Construction}

\begin{figure}[!htbp]
    \centering
    \includegraphics[width=\linewidth]{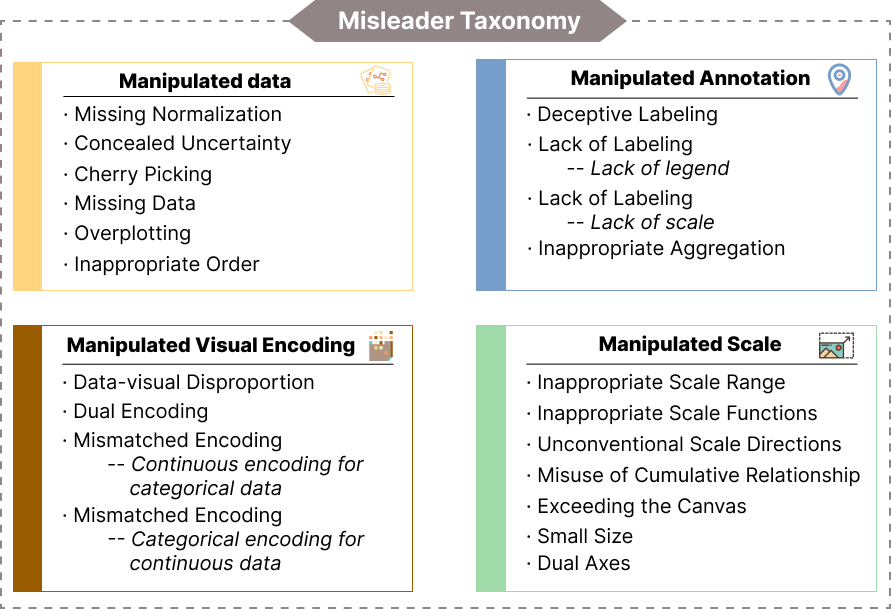}
    \caption{The taxonomy categorizes 21  misleaders into four groups based on manipulation techniques.}
    \label{fig:misleaderTaxonomy}
\end{figure}

To capture the diverse ways visualizations can mislead, we constructed a Misleader Taxonomy by consolidating deceptive strategies from academic literature and three publicly available collections of real-world misleading visualizations~\cite{lo2022misinformed, borner2019data, lan2024came}. Four data visualization experts—two postdoctoral researchers and two senior PhD students—independently reviewed these sources to compile an initial list of common misleaders. Through collaborative refinement, they merged overlapping items, clarified ambiguous definitions, and removed overly narrow cases, resulting in 21 distinct misleader types. The experts then mapped relevant chart types to each misleader, focusing on contexts where these deceptive patterns frequently occur. This process yielded 10 unique chart types and 60 distinct (misleader, chart type) pairings, ensuring broad and representative coverage. Detailed definitions and chart mappings are provided in \cref{fig:misleaderDefinition}. Finally, the misleaders were organized into a structured taxonomy (\cref{fig:misleaderTaxonomy}), forming the foundation for subsequent data augmentation.

\subsection{Seed Multiple-Choice Question Design}
Building on our Misleader Taxonomy and the 60 (misleader, chart type) pairs, we collaborated with four experts to construct a comprehensive set of ``seed MCQs'', ensuring coverage of all pairings with multiple examples per pair. This seed set was derived from two primary sources. First, experts manually reviewed MCQs from prior studies~\cite{lee2016vlat, cui2023adaptive, ge2023calvi}, identifying those that aligned with our taxonomy and pairing scheme. An MCQ was selected if at least three out of four experts agreed it was a good match for a specific (misleader, chart type) pair. This process yielded 122 MCQs covering 49 of the 60 pairs.

For the remaining 11 uncovered pairs, each expert independently crafted new misleading chart QA items, which were then refined and finalized through multiple rounds of collaborative discussion. This led to an additional 33 MCQs. In total, we compiled 155 seed MCQs, ensuring that each (misleader, chart type) pairing is represented by 2–3 well-defined examples. 

\begin{figure*}[!ht]
    \centering
    \includegraphics[width=\linewidth]{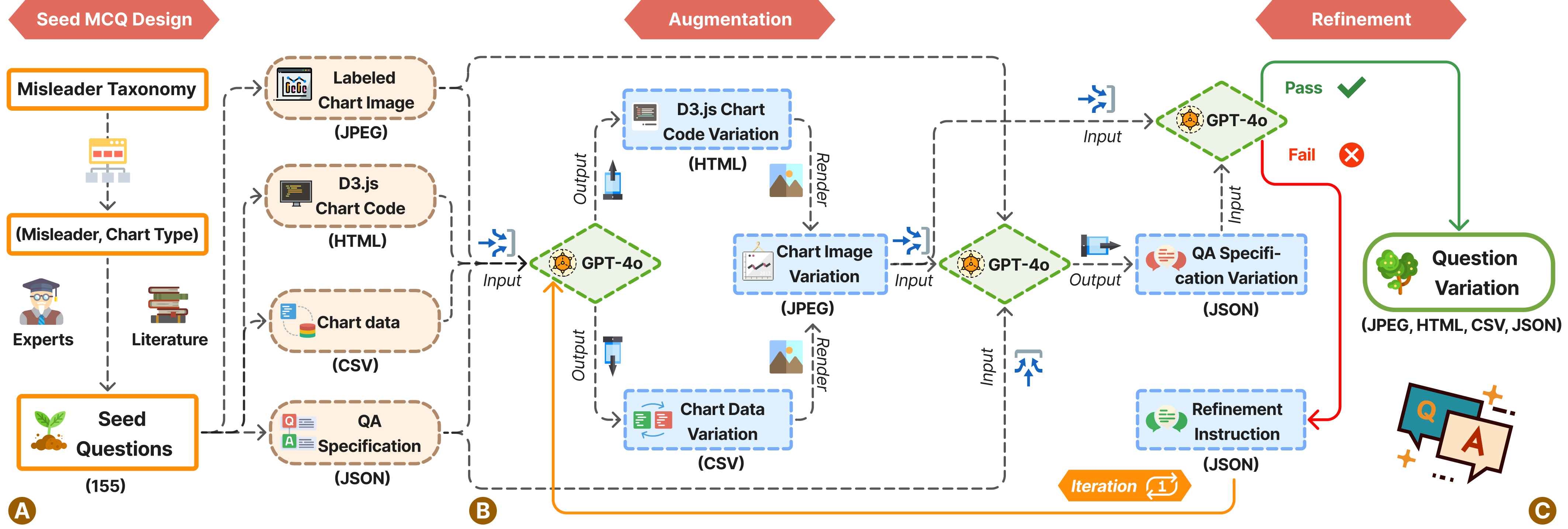}
    \caption{Overview of the Automated MCQ Augmentation and Iterative Refinement workflow. (A) \textit{Seed MCQ Design:} Questions are authored by experts, guided by the proposed misleader taxonomy and relevant literature.  (B)  \textit{Chart Variation:} MLLM modifies chart code and data to generate variations while preserving the intended misleader. (C) \textit{MCQ Augmentation and Refinement Loop}. A separate MLLM generates QA pairs and explanations, followed by an evaluation and revision loop to improve failed cases. Final outputs include variations in JPEG, HTML, CSV, and JSON.}
    \label{fig:generationPipeline}
\end{figure*}

As shown in~\cref{fig:example}, each seed MCQ includes: (1) a misleading chart, (2) a corresponding question, (3) multiple answer choices, (4) labeled correct and misleading answers, and (5) metadata with an explanation of the misleading aspect. Once finalized, all seed MCQs were encoded in a standardized format to support systematic chart and data variation. Each encoded MCQ consists of:


 \textbf{Misleading Chart Code Implementation.}  
To enable flexible generation and variation of misleading chart visualizations, each seed chart was implemented using D3.js~\cite{bostock2011d3}, a JavaScript library for highly customizable visualizations. The code was structured in modular HTML files for easy rendering, consistent coding style, and efficient generation of visual variations.

\textbf{CSV Data and JSON QA Specification.}  
Each chart was paired with a curated CSV dataset designed to reflect the associated misleader scenario. For instance, a scatter plot labeled as \textit{Cherry Picking} may use a selectively filtered dataset to exaggerate a trend (e.g., \cref{Manipulated_Data}). Corresponding MCQs were encoded in JSON format, including question text, answer choices, correct and misleading answers, and detailed metadata for compatibility and downstream processing.

 \textbf{Chart Figure Generation.}  
We rendered each chart using the implemented code and data, and developed a labeling tool (\cref{fig:labellingTool}) for experts to annotate misleading regions using bounding boxes. Both raw and annotated chart images were exported in standardized JPEG format with consistent dimensions to support scalable dataset expansion.

\subsection{MCQs Augmentation and Refinement}


Using seed MCQs for each misleader–chart type pair, we conduct a data augmentation process, leveraging general world knowledge from MLLMs (e.g., GPT-4o) to generate diverse MCQ variations while preserving the core misleading features. 

Specifically, we apply controlled perturbations to chart code and introduce randomized yet plausible variations to the CSV data. This process does not rely on the model’s training data, proprietary knowledge, or internal mechanisms, but instead uses only its general reasoning ability. By design, it minimizes the risk of model bias or knowledge leakage, ensuring that augmented examples for later experiments reflect generic reasoning rather than model-specific heuristics. The next section outlines the workflow structure, with detailed prompt templates in~\cref{prompt1}.

For each seed question, the annotated chart image, code, data, and JSON QA specification serve as core inputs to our MLLM-powered augmentation pipeline. We use ChatGPT-4o for its strong performance and efficiency, while strictly limiting its role to general-purpose tasks such as modifying HTML object attributes (e.g., color, axis scale, label position) and introducing plausible random adjustments to CSV data. These actions rely solely on general world knowledge and do not require any model-specific internal training data. The augmentation process consists of two main stages—\textit{Chart Variation} and \textit{QA Generation}—followed by an \textit{Automated Evaluation, Feedback, and Refinement Loop} to ensure high-quality outputs.



\textbf{Chart Variation:} In the first stage (\cref{fig:generationPipeline}-A), we apply controlled modifications to the chart code and underlying dataset to generate visual and contextual diversity. Specifically, the MLLM perturbs the seed D3.js code by adjusting general HTML attributes such as color schemes, axis layout, font size, or chart titles—tasks based on common web development conventions. Simultaneously, the associated CSV data is modified through random perturbations of numeric values and category labels, while maintaining the overall distribution and preserving the intended misleading effect. This stage ensures that each variation preserves the original misleader but presents it in a new surface form suitable for robust model benchmarking.

\textbf{QA Generation:} Once the chart and dataset are modified, the pipeline (\cref{fig:generationPipeline}-B) launches a local server to render the updated chart and capture it as an image. This image, along with the original seed QA specification and metadata, is then passed to another MLLM module, which adjusts the MCQ to align with the new chart while preserving the original misleading logic.


\textbf{Automated Evaluation, Feedback, and Refinement Loop:} To ensure quality and reduce manual effort in the final review stage, each generated QA pair undergoes an automated, iterative first-pass check and revision process using an MLLM module. This module assesses whether the question, chart, and answers are logically coherent and whether the intended misleader is accurately preserved. If issues such as erroneous charts, ambiguous questions, or visual-question mismatches are detected, the system provides targeted revision instructions. These revisions are fed back into the generation module in a loop that continues until the output passes all checks. By filtering and correcting obvious errors early, this process significantly reduces the burden on human reviewers. At the end of this automated stage, a total of 4,263 augmented QA samples were generated across all misleader–chart type combinations, ready for subsequent expert validation.

\begin{table*}[!ht]

\scriptsize
\centering
    \resizebox{\textwidth}{!}{ 
    \begin{tabular}{l|ccc|ccc|ccc}
    \toprule

     \multirow{2}{*}{\textbf{Model}} & \multicolumn{3}{c|}{\textsc{\textbf{Baseline}}}  & \multicolumn{3}{c|}{\textsc{\textbf{Zero-shot CoT}}}  & \multicolumn{3}{c|}{\textsc{\textbf{Pipeline}}}   \\
             
       & \textbf{W. O.} & \textbf{W. M.} & \textbf{Acc.} 
      & \textbf{W. O.} & \textbf{W. M.} & \textbf{Acc.} 
      & \textbf{W. O.} & \textbf{W. M.} & \textbf{Acc.} 
      \\
     
     \midrule

     \textsc{Random Guess} & 50.00 & 25.00 & 25.00& 50.00 & 25.00 & 25.00& 50.00 & 25.00 & 25.00 \\
     \rowcolor{HighlightBlue}
      Average (Overall) & 27.38 & 35.02 & 37.60 & 28.35 & 34.51 & 37.14 & 26.82 & 33.43 & 39.76\\
     \midrule
     \multicolumn{10}{c}{\textsc{Closed-source}} \\
     \midrule
  GPT-4o & 26.60 & 38.47 & \underline{\textbf{34.93}} & 25.57 & 37.79 & \underline{\textbf{36.64}}& 27.74 & 33.22 & 39.04\\
  GPT-4.1 & 21.92 & 43.15 & 34.93 & 19.86 & 44.29 & 35.84 & 22.60 & 37.21 & 40.18\\
  GPT-o1 & 30.02 & 35.62 & \underline{\textbf{\textcolor{HighlightRed}{34.36}}} & 24.43 & 37.44 & \underline{\textbf{\textcolor{HighlightRed}{38.13}}}& 23.29 & 34.02 & \underline{\textbf{\textcolor{HighlightRed}{42.69}}}\\
  GPT-o3 & 23.29 & 39.95 & \underline{\textbf{36.76}} & 26.94 & 39.95 & \underline{\textbf{33.11}} & 23.06 & 34.93 & 42.01\\
  GPT-o4-mini & 22.60 & 39.95 & \underline{\textbf{37.44}} & 24.43 & 39.95 & \underline{\textbf{35.62}} & 25.11 & 36.07 & 38.81\\
  Claude-3.5-Sonnet & 36.30 & 29.57 & 34.13 & 27.63 & 35.38 & 36.99 & 25.80 & 35.96 & 38.24\\
  Claude-3.7-Sonnet & 35.16 & 30.59 & 34.25 & 27.63 & 34.59 & 37.78 & 37.21 & 37.78 & 25.01\\
   Gemini-2.0-Flash & 43.49 & 25.46 & 31.05& 47.03 & 18.04 & 34.93& 42.58 & 20.78 & 36.64 \\
  Gemini-2.5-Flash & 43.15 & 18.95 & \underline{\textbf{37.90}} & 39.50 & 20.09 & \underline{\textbf{40.41}} & 37.44 & 25.11 & 37.44\\
            \arrayrulecolor{black!60}\specialrule{0.5pt}{0.5pt}{0.5pt}\arrayrulecolor{black}
            \rowcolor{HighlightBlue}
   Average (Closed-Source) & 31.39 & 33.52 & 35.08 & 29.22 & 34.17 & 36.61 & 29.43 & 32.79 & 37.78\\

     \midrule
          \multicolumn{10}{c}{\textsc{Open-source}} \\
    \midrule
    \addlinespace[2pt] 


       DeepSeek-VL2-Tiny & 28.54 & 40.52 & \underline{\textbf{30.94}} & 32.88 & 37.90 & \underline{\textbf{29.22}}& 31.74 & 35.27 & 32.99\\
  DeepSeek-VL2-Small & 26.60 & 43.61 & \underline{\textbf{29.79}} & 34.70 & 44.06 & \underline{\textbf{21.24}}& 27.40 & 43.15 & 29.45\\
  DeepSeek-VL2 & 26.48 & 43.61 & 29.91 & 30.37 & 34.70 & 34.93 & 24.43 & 38.58 & 36.99\\
   Qwen2.5-VL-3B & 35.16 & 30.60 & \underline{\textbf{34.24}} & 36.99 & 29.22 & \underline{\textbf{33.79}}& 34.70 & 27.63 & 37.67\\
    Qwen2.5-VL-7B & 27.40 & 34.93 & 37.67 & 29.22 & 33.11 & 37.67& 27.63 & 31.74 & 40.64\\
     Qwen2.5-VL-72B & 29.45 & 29.45 & \underline{\textbf{41.10}} & 28.77 & 28.77 & \underline{\textbf{42.47}}& 31.51 & 25.11 & 43.38\\
        InternVL2.5-4B-MPO & 24.20 & 39.73 & 36.07 &28.77 & 33.33 & 37.90 & 26.48 & 36.07 & 37.44\\
         InternVL2.5-8B-MPO & 19.86 & 38.36 & 41.78 & 22.61 & 34.70 & 42.69& 18.72 & 36.53 & 44.75\\
          InternVL2.5-26B-MPO & 20.78 & 36.76 & 42.47 & 29.22 & 29.68 & 41.10& 18.49 & 38.81 & 42.69\\
         InternVL2.5-78B-MPO & 20.09 & 31.96 & 47.95 & 16.89 & 36.76 & 46.35 & 18.95 & 32.31 & 48.74\\
         InternVL3-8B-MPO & 26.48 & 31.51 & 42.01 & 33.56 & 37.79 & 28.65 & 25.57 & 30.59 & 43.84\\
         InternVL3-38B-MPO & 17.81 & 34.47 & 47.72 & 19.18 & 39.50 & 41.32 & 20.78 & 35.16 & 44.06\\
          InternVL3-78B-MPO & 16.89 & 33.11 & \underline{\textbf{\textcolor{HighlightRed}{50.00}}} & 17.48 & 32.19 & \underline{\textbf{\textcolor{HighlightRed}{50.23}}} & 18.72 & 29.34 & \underline{\textbf{\textcolor{HighlightRed}{51.94}}}\\
          \arrayrulecolor{black!60}\specialrule{0.5pt}{0.5pt}{0.5pt}\arrayrulecolor{black}
          \rowcolor{HighlightBlue}
          Average (Open-Source) & 24.60 & 36.05 & 39.36 & 27.74 & 34.75 & 37.50 & 25.01 & 33.87 & 41.12\\

    \bottomrule

    \end{tabular}
    }
    \caption{Overall evaluation results of different MLLMs on Misleading ChartQA across three methods: Baseline, zero-shot CoT, and our proposed Pipeline (\cref{pipeline}). \textbf{\textit{W.O.}} refers to errors from general distractors, \textbf{\textit{W.M.}} from the misleading distractor, and \textbf{\textit{Acc.}} denotes accuracy (selection of the correct answer). Prompt templates are detailed in~\cref{prompt2,prompt3}.}
    \label{tab:summary_result}
    \vskip -0.2in

\end{table*}

\subsection{Intensive Expert Validation}

While automation filters low-quality outputs, expert validation remains crucial to ensure each augmented MCQ meets high standards. Due to the nuance of misleading charts, this stage requires intensive expert effort and cannot be reliably delegated to crowd-sourced or general annotators.

To this end, we recruited 20 PhD students specializing in data visualization—individuals with deep expertise in chart design, cognitive perception, and visual literacy—specifically to handle the complex reasoning required to evaluate misleading visual content. Each expert was compensated at $\$30$ USD per hour and followed a three-stage evaluation process using our custom annotation tool (\cref{fig:labellingTool}). This process involved verifying whether the chart reflects the intended misleader, assessing the clarity and validity of the chart and QA pair, and deciding whether to reject, revise, or approve each sample (\cref{guideline}). 

Of the 4,263 augmented QA samples, $29.02\%$ were discarded due to misalignment or irreparable chart issues, $60.52\%$ were revised by updating QA content, explanations, or making minor adjustments to chart code, and $10.46\%$ were approved without modification. Each approved sample was reviewed by two experts, and all revised samples underwent an additional check. The final dataset comprises 3,026 MCQs with corresponding charts, data, QA specifications, and misleader annotations, of which about $30\%$ received an additional validation round and are designated as a high quality testing set\footnote{https://github.com/CinderD/MisleadingChartQA}. A detailed dataset breakdown and benchmark comparison are provided in~\cref{tab:comparison}.

\section{Experiments}



In this section, we first describe our experimental setup (\cref{sec:section-experiments-subsection-setup}), followed by a comprehensive evaluation results on the Misleading ChartQA benchmark (\cref{sec:section-experiments-subsection-results}). Full implementation details are provided in the \cref{Implementation_Details}.



\subsection{Experimental Setup}
\label{sec:section-experiments-subsection-setup}




To comprehensively evaluate model performance on the Misleading ChartQA benchmark, we cover most recent widely used MLLMs, spanning both closed-source \texttt{GPT} series (4o, 4.1, o1, o3, o4-mini)~\cite{openai20244o,openai2024o1}, \texttt{Claude} series (3.5 \& 3.7 Sonnet)~\cite{anthropic2024,anthropic2025}, and \texttt{Gemini} series (2.0 \& 2.5 Flash)~\cite{deepmind2024, deepmind2025}, as well as open-sourced \texttt{DeepSeek-VL2}~\cite{wu2024deepseek}, \texttt{Qwen2.5-VL}~\cite{bai2025qwen2}, and \texttt{InternVL2.5 \& InternVL3}~\cite{chen2024expanding}, with parameter sizes ranging from 2B to 78B.


For each model, we adopt the default prompting configurations from their respective papers or official documentation as the baseline~\cite{chen2024expanding, deeplearningai-2025}. We additionally apply the zero-shot Chain-of-Thought (CoT) prompting strategy~\cite{kim2023cot} to examine how prompting affects performance on misleading questions. Finally, we compare both settings with our proposed Region-Aware Misleader Reasoning approach (referred to as \textit{Pipeline}, detailed in~\cref{pipeline}) to demonstrate its effectiveness.









\subsection{Main Results}
\label{sec:section-experiments-subsection-results}
The overall results are presented in \cref{tab:summary_result}, from which we can make the following observations:

(1)\label{conclusion1} \textbf{The Misleading ChartQA task is highly challenging}, with most models scoring around $40\%$, while even the best model reaches only about $50\%$. This contrasts sharply with other chart-related benchmarks, where state-of-the-art models typically score around $90\%$. Notably, prior research similar performance from the general public on misleading chart comprehension tests, averaging $39\%$ (SD = $16\%$)~\cite{ge2023calvi}. These findings suggest that current MLLMs, trained primarily on general corpora, perform comparably to humans and lack sufficient exposure to misleading charts—underscoring the need for a dedicated corpus and further research on this task.

\begin{figure}[!t]
    \centering
    \includegraphics[width=\linewidth]{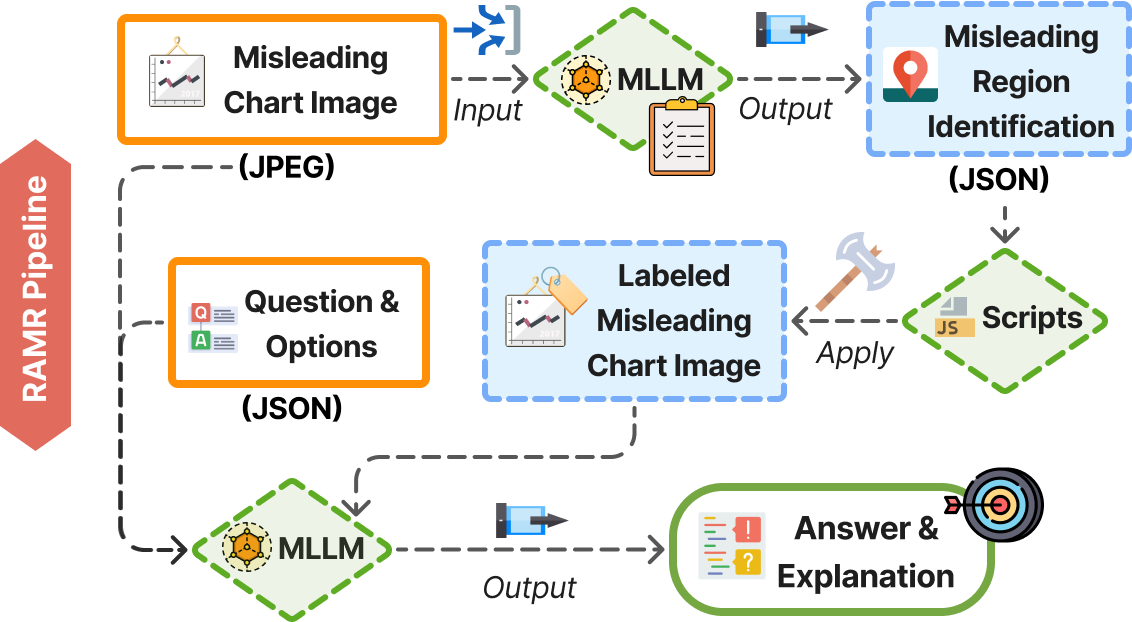}
    \caption{The Region-Aware Misleader Reasoning (RAMR) pipeline guides MLLMs to localize misleading regions first and generate answers using both original and labeled chart inputs.}

    \label{fig:pipeline}
\end{figure}


(2)\label{conclusion2} \textbf{MLLMs Are More Likely to Be Misled Than Distracted by Regular Distractors.}
Across all settings, MLLMs are more prone to selecting misleading distractors (\textit{W.M.}) than generic ones (\textit{W.O.}), despite the 2:1 ratio favoring \textit{W.O.} in random guessing. Under the baseline, \textit{W.M.} averages $36.05\%$ (open-source) and $33.52\%$ (closed-source), notably exceeding the \textit{W.O.} rates of $24.60\%$ and $31.39\%$, respectively. This pattern persists across CoT and \textit{Pipeline} settings. Even the lowest \textit{W.M.} ($32.79\%$ in closed-source \textit{Pipeline}) remains high. These results suggest MLLMs can ignore irrelevant options but still struggle to recognize and reason through deceptive chart cues, revealing a core weakness in visual critical reasoning.

(3)\label{conclusion3} \textbf{Open-Source MLLMs Surpass Closed-Source Models on Misleading Charts.}
Open-source models consistently surpass closed-source ones across all settings. In the baseline, they average $39.36\%$ accuracy versus $35.08\%$ for closed-source models—a trend that holds under both CoT and Pipeline settings. Most notably, InternVL3-78B-MPO achieves the highest scores across all settings: $50.00\%$ (Baseline), $50.23\%$ (CoT), and $51.94\%$ (Pipeline), significantly outperforming all closed-source models (with o1 \& Gemini-2.5 as the top performers). These results underscore the growing strength of open-source MLLMs in nuanced visual reasoning under large-scale parameters.


(4)\label{conclusion4} \textbf{Impact of Chain of Thought (CoT) Reasoning.} To align with prior benchmarks~\cite{kim2023cot, deeplearningai-2025, chen2024expanding}, we adopt a zero-shot CoT setting. It yields gains for most closed-source models (e.g., GPT-4o: $34.93\%\rightarrow36.64\%$, Gemini-2.5-Flash: $37.90\%\rightarrow40.41\%$), except for o3 and o4-mini—likely due to their already strong inherent reasoning abilities. In contrast, open-source models show limited or even negative effects: small and mid-sized models (e.g., DeepSeek-VL2-Tiny/Small, Qwen2.5-VL-3B) exhibit performance drops, while larger models (e.g., InternVL3-78B, Qwen2.5-VL-72B) gain only $0.5$-$1\%$. These results indicate that while CoT brings modest gains in some cases, it remains insufficient for handling misleading visual elements—especially in open-source models—highlighting the need for strategies that explicitly guide attention to deceptive features.



\subsection{Region-Aware Misleader Reasoning}\label{pipeline}


To enhance MLLMs' performance on Misleading ChartQA, we propose a multi-stage pipeline called Region-Aware Misleader Reasoning, inspired by how domain experts examine deceptive visualizations. This approach first identifies deceptive chart elements only, incorporating external scripts to assist this step-by-step process.

As illustrated in~\cref{fig:pipeline}, the pipeline begins with an MLLM independently analyzing the chart using a misleader checklist and outputting a JSON file with the coordinates and explanations of suspected misleading regions. This output is then passed to a JavaScript script that overlays bounding boxes onto the original chart. In the second stage, both the labeled chart (with explanations) and the original chart, along with the question and options, are provided to another MLLM to generate the final answer. By including both chart versions, we improve robustness to mislabeling, treating the labeled chart as a reference rather than absolute ground truth.

As shown in~\cref{tab:summary_result}-\textit{Pipeline} and discussed in~\cref{conclusion2,conclusion3}, our method consistently outperforms both baseline and zero-shot CoT settings across model families. Notably, it boosts the best closed-source model (GPT-o1) to $42.69\%$ and the best open-source model (InternVL3-78B-MPO) to $51.94\%$. Prompt templates are detailed in~\cref{prompt2,prompt3}.

\section{Discussion}

\begin{table*}[!t]
\scriptsize
\centering
\resizebox{\textwidth}{!}{ 
    \begin{tabular}{|l|c|c|c|c|}
    \toprule
    
       \multicolumn{2}{|c|}{\textbf{Misleader}} & \textbf{Wrong due to Others} & \textbf{Wrong due to Misleader} & \textbf{Accuracy} \\

     \midrule

     \midrule

  \midrule
  \multirow{8}{*}{\parbox{2cm}{\centering \textsc{Manipulated Scale}}} 
& Misuse of Cumulative Relationship & 15.24 & 48.57 & 36.19 \\
    & Small Size & 28.81 & 24.76 & 46.43 \\
     & Dual Axes & 31.27 & 35.65 & 33.08 \\
          & Exceeding the Canvas & 32.46 & 29.23 & 38.31 \\
           & Unconventional Scale Directions & 11.62 & 62.96 & 25.42 \\
  &  Inappropriate Scale Range & 25.57 & 40.29 & 32.14 \\
 &  Inappropriate Scale Functions & 28.58 & 27.29 & 44.13 \\

 \rowcolor{HighlightBlue}
& \textbf{Category Overall} & 
\underline{\textbf{\textcolor{HighlightRed}{24.79}}} & \underline{\textbf{\textcolor{HighlightRed}{38.39}}} & \underline{\textbf{\textcolor{HighlightRed}{36.53}}}\\
     \midrule
     \multirow{5}{*}{\parbox{2cm}{\centering \textsc{Manipulated Annotation}}} 
     & Deceptive Labeling & 20.24 & 26.43 & 53.81 \\
 & Lack of Labeling\textsubscript{ Lack of legend}
 & 14.64 & 35.71 & 49.64 \\
 & Lack of Labeling\textsubscript{ Lack of scales} & 34.05 & 39.76 & 26.19 \\
  & Inappropriate Aggregation & 28.43 & 41.14 & 30.43 \\
& \textbf{Category Overall} & 24.34 & 35.76 & 40.02\\
    \midrule
     \midrule
     
  \multirow{5}{*}{\parbox{2cm}{\centering \textsc{Manipulated Visual Encoding}}} 
   & Dual Encoding & 22.38 & 23.10 & 54.52 \\
  & Data-visual Disproportion & 29.46 & 37.50 & 33.04 \\

  & Mismatched Encoding\textsubscript{ Continuous encoding} & 27.62 & 22.86 & 49.52 \\
  & Mismatched Encoding\textsubscript{ Categorical encoding} & 28.66 & 27.17 & 44.17 \\
  \rowcolor{HighlightBlue}
& \textbf{Category Overall} & \underline{\textbf{\textcolor{HighlightRed}{27.03}}} & \underline{\textbf{\textcolor{HighlightRed}{27.66}}} & \underline{\textbf{\textcolor{HighlightRed}{45.31}}}\\

   \midrule

\multirow{7}{*}{\parbox{2cm}{\centering \textsc{Manipulated Data}}} 

 & Cherry Picking & 12.86 & 29.29 & 57.86 \\
  & Missing Data & 15.71 & 58.57 & 25.71 \\
  & Overplotting & 47.14 & 26.43 & 26.43 \\
  & Inappropriate Order & 30.83 & 45.60 & 23.57 \\
  & Missing Normalization & 15.71 &22.14 & 62.14 \\
 & Concealed Uncertainty & 25.00 & 25.24 & 49.76\\
& \textbf{Category Overall} & 24.54 & 34.55 & 40.91\\
     \midrule

    \bottomrule
    \end{tabular}
}
  \caption{Summary statistics for different misleader categories and types, showing average rates of \textit{Wrong due to Others}, \textit{Wrong due to Misleader}, and overall accuracy.}  
    \label{tab:misleader_statistics}
    \vskip -0.2in

\end{table*}

To further understand the limitations of current MLLMs, we present a diagnostic analysis of misleader types and chart structures, examine common failure cases, and discuss human baselines along with the potential of fine-tuning.

\subsection{Performance Across Misleader Types}

First, we analyzed MLLMs’ overall performance across misleader categories with a balanced testing set. As shown in~\cref{tab:misleader_statistics}, MLLMs perform poorest on the \textbf{Manipulated Scale} group, which records the lowest average \textit{Accuracy} ($36.53\%$) and the highest \textit{Wrong due to Misleader} rate ($38.39\%$). Scale manipulations such as \textit{unconventional directions}, or \textit{inappropriate ranges} demand precise quantitative reasoning beyond surface cues, leaving models especially vulnerable.

By contrast, the \textbf{Manipulated Visual Encoding} group attains the highest \textit{Accuracy} ($45.31\%$) and the lowest misled rate ($27.66\%$), indicating that MLLMs are more adept at recognizing perceptible irregularities like \textit{dual encoding} or \textit{mismatched encoding}. \textbf{Annotation} and \textbf{Data} manipulations fall in between. Overall, the results suggest that models are better at handling conspicuous visual flaws than subtle scale distortions requiring deeper quantitative inference. We hypothesize this gap reflects pretraining biases—models are tuned to align text with visible features rather than to conduct rigorous statistical reasoning. Example MCQs are provided in~\cref{Manipulated_Scale,Manipulated_Annotation,Manipulated_Data,Manipulated_Visual_Encoding}.

\subsection{Performance Across Chart Types}

\begin{figure}[!t]
    \centering
    \includegraphics[width=\linewidth]{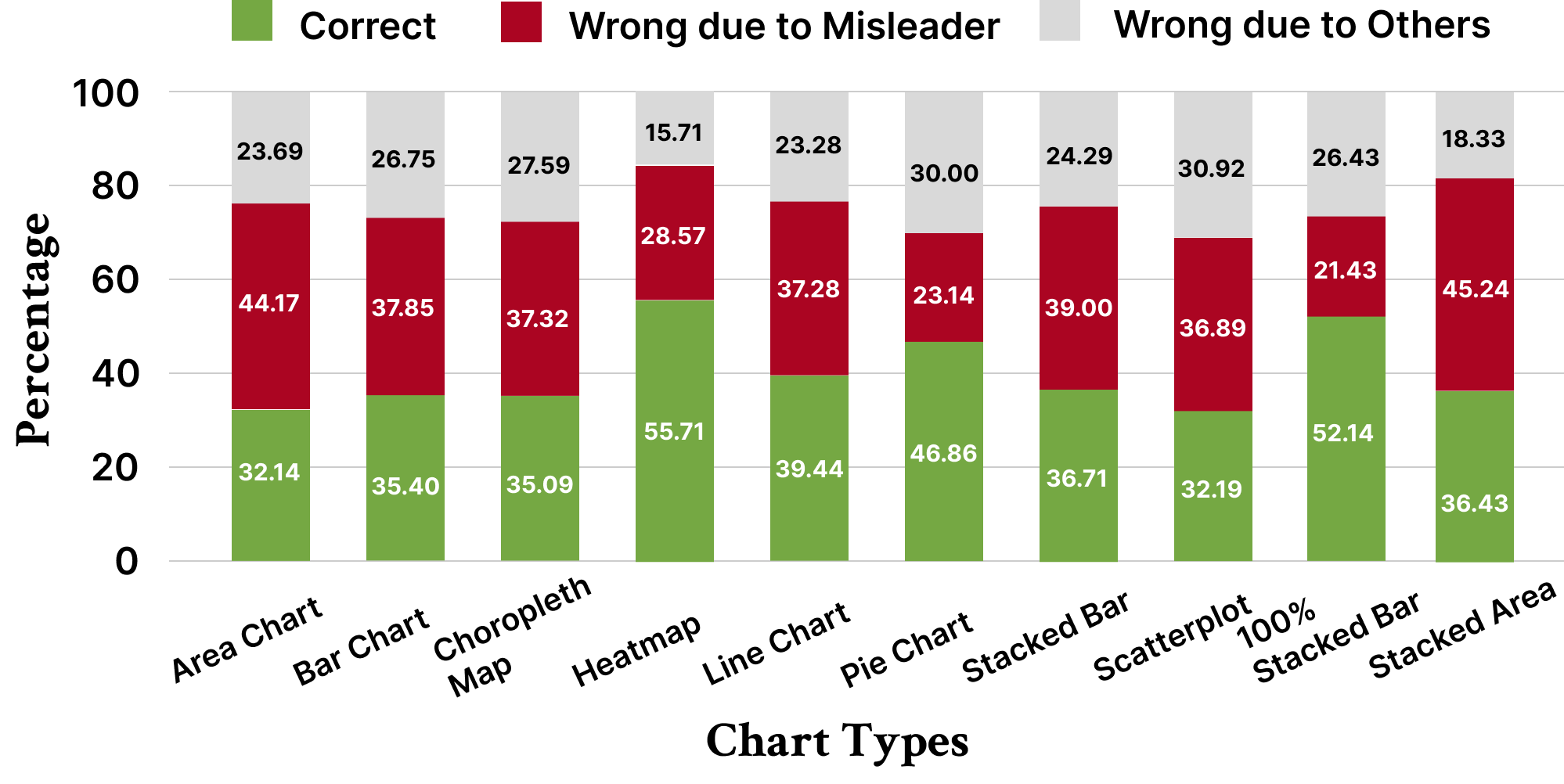}
    \caption{MLLM performance by chart type, with high misleader errors on trend/aggregation charts and more general reasoning errors on spatial formats, while normalized or grid-based charts remain more robust.}
    \label{fig:chart_type}
\end{figure}

Second, we further explored MLLMs' overall performance across different chart types. As shown in~\cref{fig:chart_type}, performance varies notably. \textbf{Heatmaps} yield the highest accuracy ($55.71\%$), followed by \textbf{100\% Stacked Bar} ($52.14\%$) and \textbf{Pie Charts} ($46.86\%$). At the low end, \textbf{Area Charts} ($32.14\%$) and \textbf{Scatterplots} ($32.19\%$) perform worst.


Error profiles reveal two dominant failure modes. (1) Trend or aggregation charts—\textbf{Area}, \textbf{Bar}, \textbf{Line}, and \textbf{Stacked Area}—show the highest \textit{Wrong due to Misleader} rates ($44.17\%$, $37.85\%$, $37.28\%$, and $45.24\%$), indicating strong susceptibility to axis and stacking manipulations. (2) Spatial or point-cloud formats (e.g., \textbf{Scatterplot} and \textbf{Choropleth Map}) exhibit elevated \textit{Wrong due to Others} ($30.92\%$ and $27.59\%$), reflecting structural/spatial reasoning errors even without explicit misleaders. By contrast, normalized or grid-structured displays like \textbf{Heatmap}, \textbf{100\% Stacked Bar}, and \textbf{Pie} pair above-average accuracy ($55.71\%$, $52.14\%$, $46.86\%$) with relatively low misleader rates ($28.57\%$, $21.43\%$, $23.14\%$).

\subsection{Error Analysis}  

To better understand model limitations, we analyze failure cases from the top-performing models: GPT-o1 and InternVL3-78B-MPO, under the proposed pipeline. Three major error types emerge:

 \textbf{Misleading Region Localization Errors.} The majority of failures stem from incorrect localization of misleading regions, leading to flawed downstream reasoning. Future research should focus on improving both the model’s ability to identify misleading elements and its precision in generating accurate region coordinates.

 \textbf{Misleader Interpretation and Reasoning Errors.} In some cases, the model correctly identifies the misleading region but fails to reason through its implications—such as recognizing a manipulated data order but not mentally reordering the data to recover the true trend. This suggests that accurate answer selection often requires not just detection of the misleader, but also corrective reasoning to reconstruct the intended information.

 \textbf{Question Misunderstanding.} A smaller subset of errors arises from misinterpreting question intent, especially involving subtle qualifiers or conditional logic—such as confusing when to choose ``Cannot be determined'' versus directly answering ``No''. This suggests future work should go beyond evaluating option selection and include more fine-grained annotation of model reasoning, particularly in tasks like Misleading ChartQA where interpretive reasoning is central.

\subsection{Human Context and Potential of Fine-Tuning}

Our benchmark does not yet include experiments with human participants to establish a direct baseline. Nevertheless, a portion of the seed questions is adapted from two standardized chart-literacy tests, where prior studies report an average public accuracy of about $39\%$~\cite{ge2023calvi,lee2016vlat}. The overall accuracy of state-of-the-art MLLMs on our benchmark falls within a similar range, suggesting that the benchmark reflects challenges comparable to those faced by general audiences, reflect the potential influence of large-scale real-world training data that shapes current MLLM performance.

At the same time, our preliminary experiments indicate that fine-tuning can provide further performance gains. A LoRA-based adaptation of InternVL3-8B-MPO improved accuracy from $42.01\%$ to $45.43\%$, outperforming both the baseline and our lightweight pipeline. While this demonstrates the value of task-specific training, the improvement remains modest and reasoning flaws persist. More extensive fine-tuning would likely require substantially larger data and computational resources, underscoring the trade-off between accuracy gains and scalability.

\section{Related Works}
Here we summarize key related work below and provide full details in~\cref{full_rw}.

\textbf{Chart Reasoning Benchmarks.}
Prior benchmarks like ChartQA~\cite{masry2022chartqa} and PlotQA~\cite{methani2020plotqa} evaluate basic chart understanding on common chart types. Recent works expand chart coverage~\cite{han2023chartllama, xia2024chartx}, add task complexity (e.g., captioning~\cite{huang2023lvlms}, summarization~\cite{rahman2023chartsumm}). However, none explicitly focus on misleading visualizations~\cite{bharti2024chartom}.

\textbf{Misleading Visualization Studies.}
Human-centered evaluations~\cite{ge2023calvi} have identified common chart misleaders and assessed reasoning via MCQs, but their limited scale is inadequate for benchmarking MLLMs. Taxonomy-driven studies~\cite{lo2022misinformed, lan2024came} emphasize design heuristics over standardized tests.


\textbf{MLLMs and Misleading Charts.}
Recent efforts~\cite{bendeck2024empirical, tonglet2025protecting} evaluate MLLMs on small sets of human-designed misleading charts, offering limited generalizability. 
\section{Conclusions}
\label{sec:conclusions}

We present Misleading ChartQA, the first benchmark for evaluating MLLMs' ability to detect and reason about misleading chart visualizations. The dataset comprises over 3,000 curated examples across 21 misleader types and 10 chart formats. We benchmark 24 MLLMs, conduct systematic analyses, and introduce a pipeline to improves model accuracy. Our work lays a foundation for advancing MLLM-based visual misinformation detection and robust chart comprehension.


\section{Limitations}
\label{sec:limitations}



\textbf{Limited Visual Prompt Design and Comparison}
In line with the original models publishers' approaches (e.g., Qwen, DeepSeek, and InternVL series), which primarily use zero-shot methods for ChartQA benchmark testing, our evaluation also adopts a zero-shot approach. While this alignment facilitates comparison, it is likely that MLLMs’ performance could be further enhanced through few-shot learning methods. Future work could explore this by incorporating few-shot techniques to potentially improve the models’ capabilities in handling misleading chart detection tasks.

\textbf{Limited Fine-Tuning Experiments}  
While we conducted a preliminary LoRA fine-tuning on InternVL3-8B-MPO and observed modest gains, our study lacks large-scale fine-tuning across different models. Due to resource constraints, we were unable to explore full fine-tuning on larger architectures such as InternVL2.5-78B-MPO. Future work should investigate broader fine-tuning strategies to more comprehensively assess the potential of model adaptation on Misleading ChartQA.
\section{Ethics Statement}
This work does not involve the collection or use of any human-related data. All materials are chart-based and either generated or derived from publicly available sources. No ethical concerns were identified in the preparation of this dataset or study.
\section*{Acknowledgments}

This project is supported by RGC GRF grant No. 16218724. The authors would like to thank Liwenhan Xie, Haobo Li, Wenshuo Zhang, Yumeng Li, Yuhang Zeng, and other members of VisLab for their valuable assistance in this work.
\balance
\bibliography{main}

\begin{thebibliography}{56}
\providecommand{\natexlab}[1]{#1}

\bibitem[{Akter et~al.(2024)Akter, Lee, Chang, Bisk, and Nyberg}]{akter2024visreas}
Syeda~Nahida Akter, Sangwu Lee, Yingshan Chang, Yonatan Bisk, and Eric Nyberg. 2024.
\newblock Visreas: Complex visual reasoning with unanswerable questions.
\newblock \emph{arXiv preprint arXiv:2403.10534}.

\bibitem[{Anthropic(2024)}]{anthropic2024}
Anthropic. 2024.
\newblock \href {https://www-cdn.anthropic.com/fed9cc193a14b84131812372d8d5857f8f304c52/Model_Card_Claude_3_Addendum.pdf} {{Claude 3.5 Sonnet Model Card Addendum}}.

\bibitem[{Anthropic(2025)}]{anthropic2025}
Anthropic. 2025.
\newblock \href {https://www.anthropic.com/claude/sonnet} {{Claude 3.7 Sonnet}}.

\bibitem[{Bai et~al.(2025)Bai, Chen, Liu, Wang, Ge, Song, Dang, Wang, Wang, Tang et~al.}]{bai2025qwen2}
Shuai Bai, Keqin Chen, Xuejing Liu, Jialin Wang, Wenbin Ge, Sibo Song, Kai Dang, Peng Wang, Shijie Wang, Jun Tang, and 1 others. 2025.
\newblock Qwen2. 5-vl technical report.
\newblock \emph{arXiv preprint arXiv:2502.13923}.

\bibitem[{Bendeck and Stasko(2024)}]{bendeck2024empirical}
Alexander Bendeck and John Stasko. 2024.
\newblock An empirical evaluation of the gpt-4 multimodal language model on visualization literacy tasks.
\newblock \emph{IEEE Transactions on Visualization and Computer Graphics}.

\bibitem[{Bharti et~al.(2024)Bharti, Cheng, Rho, Zhang, Cai, Lee, Rau, and Zhu}]{bharti2024chartom}
Shubham Bharti, Shiyun Cheng, Jihyun Rho, Jianrui Zhang, Mu~Cai, Yong~Jae Lee, Martina Rau, and Xiaojin Zhu. 2024.
\newblock Chartom: A visual theory-of-mind benchmark for multimodal large language models.
\newblock \emph{arXiv preprint arXiv:2408.14419}.

\bibitem[{B{\"o}rner et~al.(2019)B{\"o}rner, Bueckle, and Ginda}]{borner2019data}
Katy B{\"o}rner, Andreas Bueckle, and Michael Ginda. 2019.
\newblock Data visualization literacy: Definitions, conceptual frameworks, exercises, and assessments.
\newblock \emph{Proceedings of the National Academy of Sciences}, 116(6):1857--1864.

\bibitem[{Bostock et~al.(2011)Bostock, Ogievetsky, and Heer}]{bostock2011d3}
Michael Bostock, Vadim Ogievetsky, and Jeffrey Heer. 2011.
\newblock D$^3$ data-driven documents.
\newblock \emph{IEEE transactions on visualization and computer graphics}, 17(12):2301--2309.

\bibitem[{Boy et~al.(2014)Boy, Rensink, Bertini, and Fekete}]{boy2014principled}
Jeremy Boy, Ronald~A Rensink, Enrico Bertini, and Jean-Daniel Fekete. 2014.
\newblock A principled way of assessing visualization literacy.
\newblock \emph{IEEE transactions on visualization and computer graphics}, 20(12):1963--1972.

\bibitem[{Chen et~al.(2024{\natexlab{a}})Chen, Kong, Wei, Liu, Ge, Zhao, Sun, Han, and Zhang}]{chen2024onechart}
Jinyue Chen, Lingyu Kong, Haoran Wei, Chenglong Liu, Zheng Ge, Liang Zhao, Jianjian Sun, Chunrui Han, and Xiangyu Zhang. 2024{\natexlab{a}}.
\newblock Onechart: Purify the chart structural extraction via one auxiliary token.
\newblock In \emph{Proceedings of the 32nd ACM International Conference on Multimedia}, pages 147--155.

\bibitem[{Chen et~al.(2024{\natexlab{b}})Chen, Wang, Cao, Liu, Gao, Cui, Zhu, Ye, Tian, Liu et~al.}]{chen2024expanding}
Zhe Chen, Weiyun Wang, Yue Cao, Yangzhou Liu, Zhangwei Gao, Erfei Cui, Jinguo Zhu, Shenglong Ye, Hao Tian, Zhaoyang Liu, and 1 others. 2024{\natexlab{b}}.
\newblock Expanding performance boundaries of open-source multimodal models with model, data, and test-time scaling.
\newblock \emph{arXiv preprint arXiv:2412.05271}.

\bibitem[{Cheng et~al.(2023)Cheng, Dai, and Hauptmann}]{cheng2023chartreader}
Zhi-Qi Cheng, Qi~Dai, and Alexander~G Hauptmann. 2023.
\newblock Chartreader: A unified framework for chart derendering and comprehension without heuristic rules.
\newblock In \emph{Proceedings of the IEEE/CVF International Conference on Computer Vision}, pages 22202--22213.

\bibitem[{Cui et~al.(2023)Cui, Lily, Ding, Yang, Harrison, and Kay}]{cui2023adaptive}
Yuan Cui, W~Ge Lily, Yiren Ding, Fumeng Yang, Lane Harrison, and Matthew Kay. 2023.
\newblock Adaptive assessment of visualization literacy.
\newblock \emph{IEEE Transactions on Visualization and Computer Graphics}.

\bibitem[{DeepLearning.AI(2025)}]{deeplearningai-2025}
DeepLearning.AI. 2025.
\newblock \href {https://www.deeplearning.ai/short-courses/chatgpt-prompt-engineering-for-developers/} {{ChatGPT Prompt Engineering for Developers - DeepLearning.AI}}.

\bibitem[{Deepmind(2024)}]{deepmind2024}
Deepmind. 2024.
\newblock \href {https://aistudio.google.com/prompts/new_chat?model=gemini-2.0-flash-exp} {{Gemini 2.0 Flash}}.

\bibitem[{Deepmind(2025)}]{deepmind2025}
Deepmind. 2025.
\newblock \href {https://deepmind.google/technologies/gemini/flash/} {{Gemini 2.5 Flash}}.

\bibitem[{Ge et~al.(2023)Ge, Cui, and Kay}]{ge2023calvi}
Lily~W Ge, Yuan Cui, and Matthew Kay. 2023.
\newblock Calvi: Critical thinking assessment for literacy in visualizations.
\newblock In \emph{Proceedings of the 2023 CHI conference on human factors in computing systems}, pages 1--18.

\bibitem[{Han et~al.(2023)Han, Zhang, Chen, Yang, Wang, Yu, Fu, and Zhang}]{han2023chartllama}
Yucheng Han, Chi Zhang, Xin Chen, Xu~Yang, Zhibin Wang, Gang Yu, Bin Fu, and Hanwang Zhang. 2023.
\newblock Chartllama: A multimodal llm for chart understanding and generation.
\newblock \emph{arXiv preprint arXiv:2311.16483}.

\bibitem[{He et~al.(2024)He, Zhang, Jin, Yuan, Yiu et~al.}]{he2024tubench}
Xingwei He, Qianru Zhang, A~Jin, Yuan Yuan, Siu-Ming Yiu, and 1 others. 2024.
\newblock Tubench: Benchmarking large vision-language models on trustworthiness with unanswerable questions.
\newblock \emph{arXiv preprint arXiv:2410.04107}.

\bibitem[{Hong et~al.(2025)Hong, Seto, Fan, and Maciejewski}]{hong2025llms}
Jiayi Hong, Christian Seto, Arlen Fan, and Ross Maciejewski. 2025.
\newblock Do llms have visualization literacy? an evaluation on modified visualizations to test generalization in data interpretation.
\newblock \emph{IEEE Transactions on Visualization and Computer Graphics}.

\bibitem[{Huang et~al.(2023)Huang, Zhou, Chan, Fung, Wang, Zhang, Chang, and Ji}]{huang2023lvlms}
Kung-Hsiang Huang, Mingyang Zhou, Hou~Pong Chan, Yi~R Fung, Zhenhailong Wang, Lingyu Zhang, Shih-Fu Chang, and Heng Ji. 2023.
\newblock Do lvlms understand charts? analyzing and correcting factual errors in chart captioning.
\newblock \emph{arXiv preprint arXiv:2312.10160}.

\bibitem[{Huff(2023)}]{huff1954lie}
Darrell Huff. 2023.
\newblock \emph{How to lie with statistics}.
\newblock Penguin UK.

\bibitem[{Kahou et~al.(2017)Kahou, Michalski, Atkinson, K{\'a}d{\'a}r, Trischler, and Bengio}]{kahou2017figureqa}
Samira~Ebrahimi Kahou, Vincent Michalski, Adam Atkinson, {\'A}kos K{\'a}d{\'a}r, Adam Trischler, and Yoshua Bengio. 2017.
\newblock Figureqa: An annotated figure dataset for visual reasoning.
\newblock \emph{arXiv preprint arXiv:1710.07300}.

\bibitem[{Kantharaj et~al.(2022)Kantharaj, Leong, Lin, Masry, Thakkar, Hoque, and Joty}]{kantharaj2022chart}
Shankar Kantharaj, Rixie Tiffany~Ko Leong, Xiang Lin, Ahmed Masry, Megh Thakkar, Enamul Hoque, and Shafiq Joty. 2022.
\newblock Chart-to-text: A large-scale benchmark for chart summarization.
\newblock \emph{arXiv preprint arXiv:2203.06486}.

\bibitem[{Kennedy et~al.(2000)Kennedy, Kopp et~al.}]{kennedy2000understanding}
Melita Kennedy, Steve Kopp, and 1 others. 2000.
\newblock \emph{Understanding map projections}, volume~8.
\newblock Esri Redlands, CA.

\bibitem[{Kim et~al.(2023)Kim, Joo, Kim, Jang, Ye, Shin, and Seo}]{kim2023cot}
Seungone Kim, Se~June Joo, Doyoung Kim, Joel Jang, Seonghyeon Ye, Jamin Shin, and Minjoon Seo. 2023.
\newblock The cot collection: Improving zero-shot and few-shot learning of language models via chain-of-thought fine-tuning.
\newblock \emph{arXiv preprint arXiv:2305.14045}.

\bibitem[{King(1986)}]{king1986not}
Gary King. 1986.
\newblock How not to lie with statistics: Avoiding common mistakes in quantitative political science.
\newblock \emph{American Journal of Political Science}, pages 666--687.

\bibitem[{Lan and Liu(2024)}]{lan2024came}
Xingyu Lan and Yu~Liu. 2024.
\newblock “i came across a junk”: Understanding design flaws of data visualization from the public's perspective.
\newblock \emph{IEEE Transactions on Visualization and Computer Graphics}.

\bibitem[{Lauer and O'Brien(2020)}]{lauer2020deceptive}
Claire Lauer and Shaun O'Brien. 2020.
\newblock The deceptive potential of common design tactics used in data visualizations.
\newblock In \emph{Proceedings of the 38th ACM International Conference on Design of Communication}, pages 1--9.

\bibitem[{Lee et~al.(2016)Lee, Kim, and Kwon}]{lee2016vlat}
Sukwon Lee, Sung-Hee Kim, and Bum~Chul Kwon. 2016.
\newblock Vlat: Development of a visualization literacy assessment test.
\newblock \emph{IEEE transactions on visualization and computer graphics}, 23(1):551--560.

\bibitem[{Li et~al.(2025)Li, Jung, Chen, Wang, Wang, Qu, and Lau}]{li2025pipe}
Haobo Li, Eunseo Jung, Zixin Chen, Zhaowei Wang, Yueya Wang, Huamin Qu, and Alexis Kai~Hon Lau. 2025.
\newblock Pipe: Physics-informed position encoding for alignment of satellite images and time series.
\newblock \emph{arXiv preprint arXiv:2506.14786}.

\bibitem[{Li et~al.(2024)Li, Wang, Wang, Lau, and Qu}]{li2024cllmate}
Haobo Li, Zhaowei Wang, Jiachen Wang, Alexis Kai~Hon Lau, and Huamin Qu. 2024.
\newblock Cllmate: A multimodal llm for weather and climate events forecasting.
\newblock \emph{arXiv preprint arXiv:2409.19058}.

\bibitem[{Liu et~al.(2022)Liu, Piccinno, Krichene, Pang, Lee, Joshi, Altun, Collier, and Eisenschlos}]{liu2022matcha}
Fangyu Liu, Francesco Piccinno, Syrine Krichene, Chenxi Pang, Kenton Lee, Mandar Joshi, Yasemin Altun, Nigel Collier, and Julian~Martin Eisenschlos. 2022.
\newblock Matcha: Enhancing visual language pretraining with math reasoning and chart derendering.
\newblock \emph{arXiv preprint arXiv:2212.09662}.

\bibitem[{Lo et~al.(2022)Lo, Gupta, Shigyo, Wu, Bertini, and Qu}]{lo2022misinformed}
Leo Yu-Ho Lo, Ayush Gupta, Kento Shigyo, Aoyu Wu, Enrico Bertini, and Huamin Qu. 2022.
\newblock Misinformed by visualization: What do we learn from misinformative visualizations?
\newblock In \emph{Computer Graphics Forum}, volume~41, pages 515--525. Wiley Online Library.

\bibitem[{Lo and Qu(2024)}]{lo2024good}
Leo Yu-Ho Lo and Huamin Qu. 2024.
\newblock How good (or bad) are llms at detecting misleading visualizations?
\newblock \emph{IEEE Transactions on Visualization and Computer Graphics}.

\bibitem[{Masry et~al.(2022)Masry, Long, Tan, Joty, and Hoque}]{masry2022chartqa}
Ahmed Masry, Do~Xuan Long, Jia~Qing Tan, Shafiq Joty, and Enamul Hoque. 2022.
\newblock Chartqa: A benchmark for question answering about charts with visual and logical reasoning.
\newblock \emph{arXiv preprint arXiv:2203.10244}.

\bibitem[{McFall-Johnsen(2020)}]{mcfall-johnsen-2020}
Morgan McFall-Johnsen. 2020.
\newblock \href {https://www.businessinsider.com/graph-shows-georgia-bungling-coronavirus-data-2020-5} {{A 'cuckoo' graph with no sense of time or place shows how Georgia bungled coronavirus data as it reopens}}.

\bibitem[{Methani et~al.(2020)Methani, Ganguly, Khapra, and Kumar}]{methani2020plotqa}
Nitesh Methani, Pritha Ganguly, Mitesh~M Khapra, and Pratyush Kumar. 2020.
\newblock Plotqa: Reasoning over scientific plots.
\newblock In \emph{Proceedings of the IEEE/CVF Winter Conference on Applications of Computer Vision}, pages 1527--1536.

\bibitem[{Miyai et~al.(2024)Miyai, Yang, Zhang, Ming, Yu, Irie, Li, Li, Liu, and Aizawa}]{miyai2024unsolvable}
Atsuyuki Miyai, Jingkang Yang, Jingyang Zhang, Yifei Ming, Qing Yu, Go~Irie, Yixuan Li, Hai Li, Ziwei Liu, and Kiyoharu Aizawa. 2024.
\newblock Unsolvable problem detection: Evaluating trustworthiness of large multimodal models.

\bibitem[{O'Brien(2024)}]{obrien-2024}
Lotti O'Brien. 2024.
\newblock \href {https://www.express.co.uk/news/world/1958962/world-maps-mercator-projection-china-europe-africa} {{Maps of world 'completely misleading' as true size of Europe, China and Africa revealed}}.

\bibitem[{OpenAI(2024{\natexlab{a}})}]{openai20244o}
OpenAI. 2024{\natexlab{a}}.
\newblock \href {https://openai.com/index/gpt-4o-system-card/} {{Hello GPT-4o}}.

\bibitem[{OpenAI(2024{\natexlab{b}})}]{openai2024o1}
OpenAI. 2024{\natexlab{b}}.
\newblock \href {https://openai.com/o1/} {{Introducing OpenAI o1}}.

\bibitem[{Pandey et~al.(2015)Pandey, Rall, Satterthwaite, Nov, and Bertini}]{pandey2015deceptive}
Anshul~Vikram Pandey, Katharina Rall, Margaret~L Satterthwaite, Oded Nov, and Enrico Bertini. 2015.
\newblock How deceptive are deceptive visualizations? an empirical analysis of common distortion techniques.
\newblock In \emph{Proceedings of the 33rd annual acm conference on human factors in computing systems}, pages 1469--1478.

\bibitem[{Rahman et~al.(2023)Rahman, Hasan, Farhad, Laskar, Ashmafee, and Kamal}]{rahman2023chartsumm}
Raian Rahman, Rizvi Hasan, Abdullah~Al Farhad, Md~Tahmid~Rahman Laskar, Md~Hamjajul Ashmafee, and Abu Raihan~Mostofa Kamal. 2023.
\newblock Chartsumm: A comprehensive benchmark for automatic chart summarization of long and short summaries.
\newblock \emph{arXiv preprint arXiv:2304.13620}.

\bibitem[{Ray et~al.(2016)Ray, Christie, Bansal, Batra, and Parikh}]{ray2016question}
Arijit Ray, Gordon Christie, Mohit Bansal, Dhruv Batra, and Devi Parikh. 2016.
\newblock Question relevance in vqa: identifying non-visual and false-premise questions.
\newblock \emph{arXiv preprint arXiv:1606.06622}.

\bibitem[{Sun et~al.(2024)Sun, Xin, Sun, Xu, Yang, Dong, Tang, and Chen}]{sun2024large}
Yushi Sun, Hao Xin, Kai Sun, Yifan~Ethan Xu, Xiao Yang, Xin~Luna Dong, Nan Tang, and Lei Chen. 2024.
\newblock Are large language models a good replacement of taxonomies?
\newblock \emph{Proceedings of the VLDB Endowment}, 17(11):2919--2932.

\bibitem[{Tonglet et~al.(2025)Tonglet, Tuytelaars, Moens, and Gurevych}]{tonglet2025protecting}
Jonathan Tonglet, Tinne Tuytelaars, Marie-Francine Moens, and Iryna Gurevych. 2025.
\newblock Protecting multimodal large language models against misleading visualizations.
\newblock \emph{arXiv preprint arXiv:2502.20503}.

\bibitem[{Tufte and Graves-Morris(1983)}]{tufte1983visual}
Edward~R Tufte and Peter~R Graves-Morris. 1983.
\newblock \emph{The visual display of quantitative information}, volume~2.
\newblock Graphics press Cheshire, CT.

\bibitem[{Vardi et~al.(2025)Vardi, Nir, and Shamir}]{vardi2025clip}
Ben Vardi, Oron Nir, and Ariel Shamir. 2025.
\newblock Clip-up: Clip-based unanswerable problem detection for visual question answering.
\newblock \emph{arXiv preprint arXiv:2501.01371}.

\bibitem[{Wang et~al.(2024)Wang, Huey, Sheng, Mehta, and Wang}]{wang2024scidasynth}
Xingbo Wang, Samantha~L Huey, Rui Sheng, Saurabh Mehta, and Fei Wang. 2024.
\newblock Scidasynth: Interactive structured knowledge extraction and synthesis from scientific literature with large language model.
\newblock \emph{arXiv preprint arXiv:2404.13765}.

\bibitem[{Wu et~al.(2024{\natexlab{a}})Wu, Yan, Shen, Wang, Tang, and Luo}]{wu2024chartinsights}
Yifan Wu, Lutao Yan, Leixian Shen, Yunhai Wang, Nan Tang, and Yuyu Luo. 2024{\natexlab{a}}.
\newblock Chartinsights: Evaluating multimodal large language models for low-level chart question answering.
\newblock In \emph{Findings of the Association for Computational Linguistics: EMNLP 2024}, pages 12174--12200.

\bibitem[{Wu et~al.(2024{\natexlab{b}})Wu, Chen, Pan, Liu, Liu, Dai, Gao, Ma, Wu, Wang et~al.}]{wu2024deepseek}
Zhiyu Wu, Xiaokang Chen, Zizheng Pan, Xingchao Liu, Wen Liu, Damai Dai, Huazuo Gao, Yiyang Ma, Chengyue Wu, Bingxuan Wang, and 1 others. 2024{\natexlab{b}}.
\newblock Deepseek-vl2: Mixture-of-experts vision-language models for advanced multimodal understanding.
\newblock \emph{arXiv preprint arXiv:2412.10302}.

\bibitem[{Xia et~al.(2024)Xia, Zhang, Ye, Yan, Liu, Zhou, Chen, Dou, Shi, Yan et~al.}]{xia2024chartx}
Renqiu Xia, Bo~Zhang, Hancheng Ye, Xiangchao Yan, Qi~Liu, Hongbin Zhou, Zijun Chen, Min Dou, Botian Shi, Junchi Yan, and 1 others. 2024.
\newblock Chartx \& chartvlm: A versatile benchmark and foundation model for complicated chart reasoning.
\newblock \emph{arXiv preprint arXiv:2402.12185}.

\bibitem[{Xu et~al.(2023)Xu, Du, Qi, Xu, Yuan, and Guo}]{xu2023chartbench}
Zhengzhuo Xu, Sinan Du, Yiyan Qi, Chengjin Xu, Chun Yuan, and Jian Guo. 2023.
\newblock Chartbench: A benchmark for complex visual reasoning in charts.
\newblock \emph{arXiv preprint arXiv:2312.15915}.

\bibitem[{Yang et~al.(2024)Yang, Sun, Xin, Sun, Bhalla, Chen, Choudhary, Gui, Jiang, Jiang et~al.}]{yang2024crag}
Xiao Yang, Kai Sun, Hao Xin, Yushi Sun, Nikita Bhalla, Xiangsen Chen, Sajal Choudhary, Rongze Gui, Ziran Jiang, Ziyu Jiang, and 1 others. 2024.
\newblock Crag-comprehensive rag benchmark.
\newblock \emph{Advances in Neural Information Processing Systems}, 37:10470--10490.

\bibitem[{Zeng et~al.(2024)Zeng, Lin, Ye, and Zeng}]{zeng2024advancing}
Xingchen Zeng, Haichuan Lin, Yilin Ye, and Wei Zeng. 2024.
\newblock Advancing multimodal large language models in chart question answering with visualization-referenced instruction tuning.
\newblock \emph{IEEE Transactions on Visualization and Computer Graphics}.

\end{thebibliography}
\clearpage
\nobalance
\appendix

\section{Appendix}\label{appendix}
\subsection{Full Related Works}\label{full_rw}
With the rapid progress of multimodal large language models (MLLMs), a growing body of research has examined their abilities in visual reasoning, chart understanding, and robustness under challenging conditions~\cite{sun2024large,li2025pipe,li2024cllmate,wang2024scidasynth,yang2024crag}.

\subsubsection{Chart Reasoning Benchmarks}

\begin{figure*}[!ht]
    \centering
    \includegraphics[width=\textwidth]{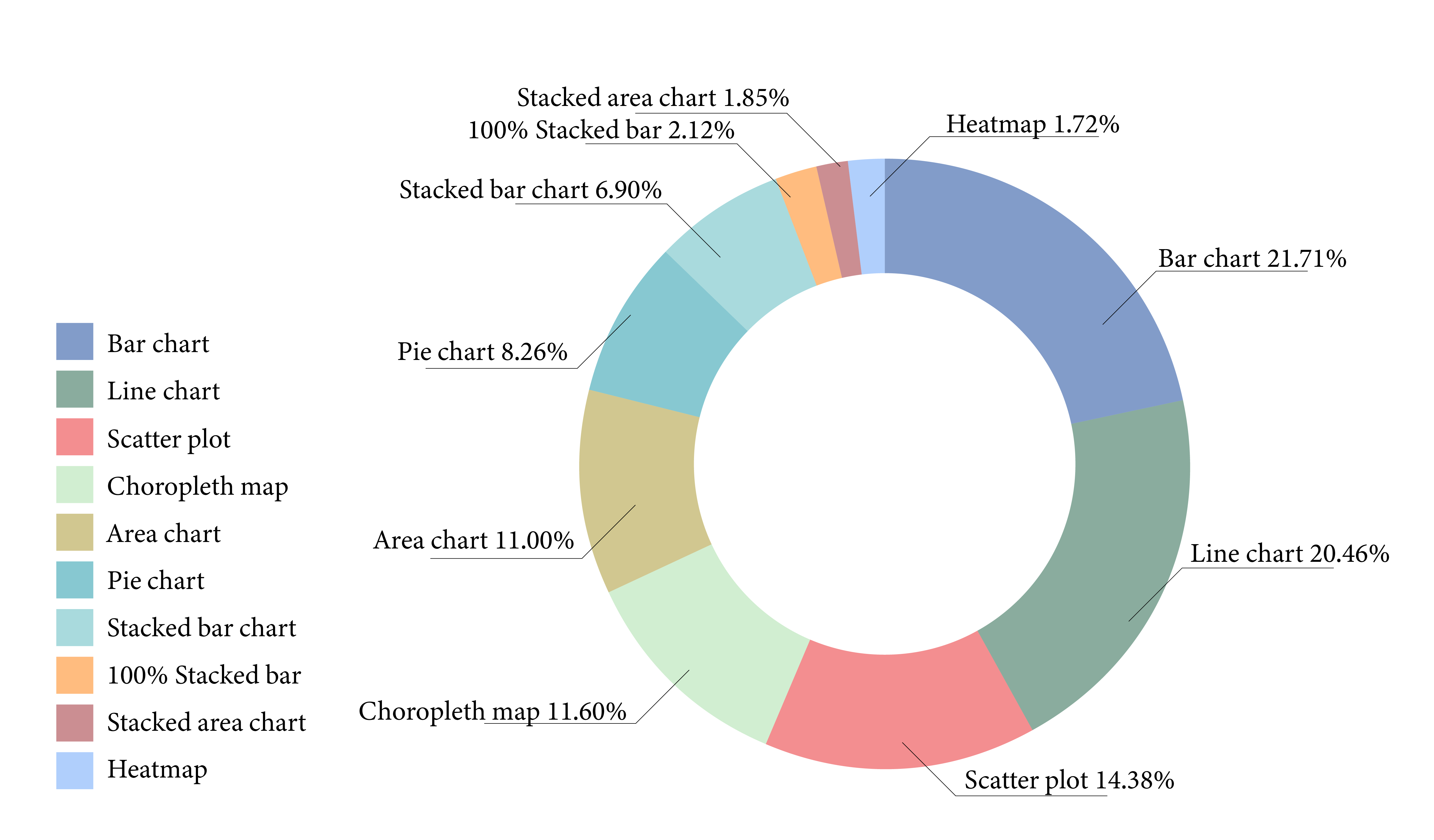}
    \caption{Breakdown of Chart Types in the Misleading ChartQA Dataset. \textbf{We intentionally balanced samples per (misleader, chart type) pair to reflect the natural mapping between chart types and supported misleaders} (e.g., heatmaps support fewer misleaders than bar charts). As a result, the overall chart distribution is uneven—mirroring real-world usage, where chart types like $100\%$ stacked bars and stacked area charts are less common than bar or line charts.}
    \label{fig:chartdistribution}
\end{figure*}

Chart Reasoning has emerged as a key area of focus within the vision-language community, with several benchmarks developed to assess models’ abilities to interpret and reason about charts. Early datasets such as ChartQA~\cite{masry2022chartqa} and PlotQA~\cite{methani2020plotqa} primarily evaluated basic chart understanding, focusing on three common chart types. These datasets were relatively straightforward for recent MLLMs to solve~\cite{li2024cllmate}. Subsequent benchmarks have either expanded chart type coverage~\cite{han2023chartllama, xia2024chartx, xu2023chartbench} or refined the complexity of tasks, distinguishing between high-level tasks (e.g., chart captioning, chart summarization~\cite{kantharaj2022chart, rahman2023chartsumm, cheng2023chartreader, huang2023lvlms,liu2022matcha}) and low-level tasks (e.g., extracting numerical values~\cite{kahou2017figureqa, wu2024chartinsights}). Some works have also introduced more complex tasks such as chart structure extraction~\cite{chen2024onechart}. A detailed comparison of chart variety with existing benchmarks is provided in~\cref{tab:comparison,fig:chartdistribution}. 

\subsubsection{Misleading Chart Visualizations}

Misleading chart visualizations have long been a significant topic in data visualization and human-computer interaction~\cite{king1986not,pandey2015deceptive,lauer2020deceptive}. Several standardized tests have been designed to evaluate human chart understanding and reasoning abilities~\cite{lee2016vlat, boy2014principled, borner2019data}. Recent efforts have evolved to emphasize critical thinking in chart comprehension, identifying around 10 categories of common misleaders in charts and formulating nuanced questions for human testing~\cite{ge2023calvi, cui2023adaptive}. However, these question sets consist of only about 40 questions, each addressing one or two examples of (misleader, chart type) combinations, which limits their effectiveness for evaluating MLLMs. Other latest studies have attempted to summarize common misleading visualization practices~\cite{lo2022misinformed, lan2024came}, but these focus on broad visualization design issues that do not directly apply to chart understanding tasks.

\subsubsection{Unanswerable Question Detection}

Prior work has studied unanswerable questions in VQA, where the challenge lies in detecting false-premise or non-visual queries and abstaining from answering~\cite{ray2016question,miyai2024unsolvable,he2024tubench,akter2024visreas,vardi2025clip}. Our setting differs in that all questions are answerable given the chart, but the visual design may intentionally mislead. Instead of abstention, models must identify deceptive encodings and still produce the correct answer, highlighting a complementary dimension of robustness in multimodal reasoning.

\subsubsection{MLLMs in Misleading Chart Comprehension}

Several recent studies have empirically evaluated MLLMs’ performance in understanding misleading chart visualizations by testing them on existing standardized tests designed for humans~\cite{bendeck2024empirical, tonglet2025protecting, hong2025llms, lo2024good, zeng2024advancing}. These studies typically involved a limited number of models and questions, making it difficult to draw reliable conclusions about MLLMs’ ability. In contrast, our work constructs a diverse benchmark with over 3,000 samples, covering a broad range of misleaders and chart types. Through a comprehensive evaluation of 16 state-of-the-art MLLMs, we establish a strong foundation for this task first-ever.
\newpage

\onecolumn
\subsection{Real-world examples: misleading charts}
\begin{figure*}[!h]
    \centering
    \includegraphics[width=\linewidth]{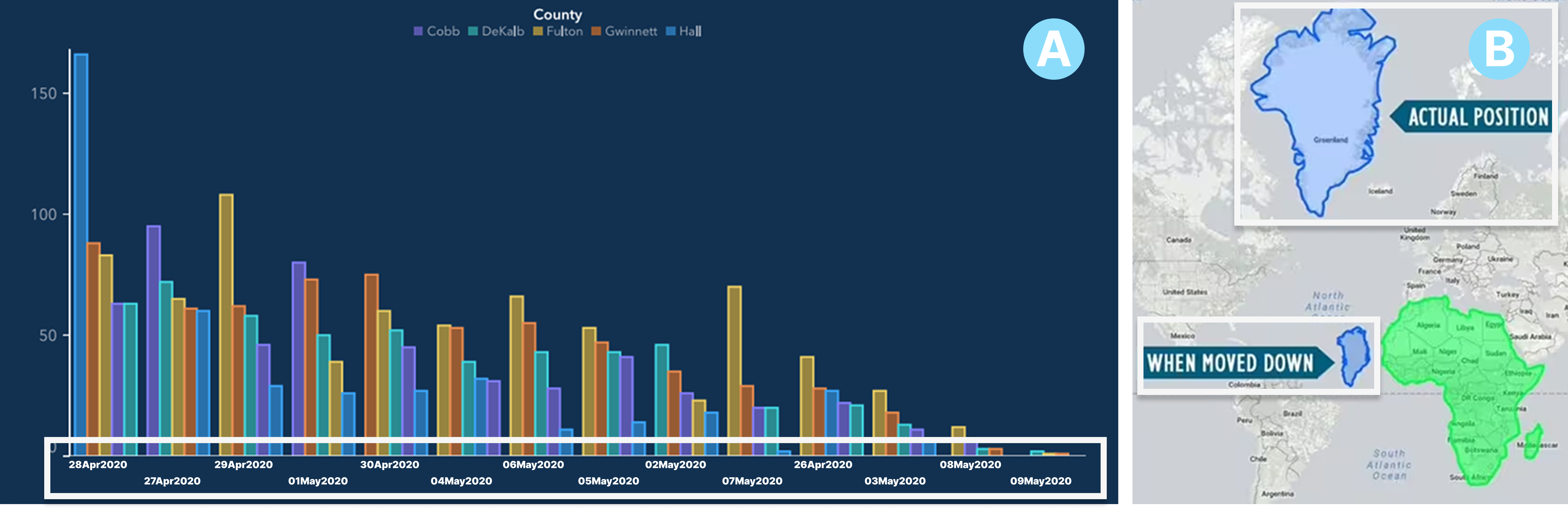}
    \caption{Two real-world examples of misleading chart visualizations. \textbf{(A)} A bar chart of COVID-19 cases across five counties, sorted by case count rather than by date, creating the false impression of a declining trend unless viewers carefully examine the x-axis. \textbf{(B)} The commonly used world map projection, which misrepresents Greenland as being the same size as Africa, despite Africa being significantly larger.}
    \label{fig:misleading_samples}
\end{figure*}

\newpage
\subsection{Misleader Definition}
\begin{figure*}[!ht]
    \centering
    \includegraphics[width=\textwidth]{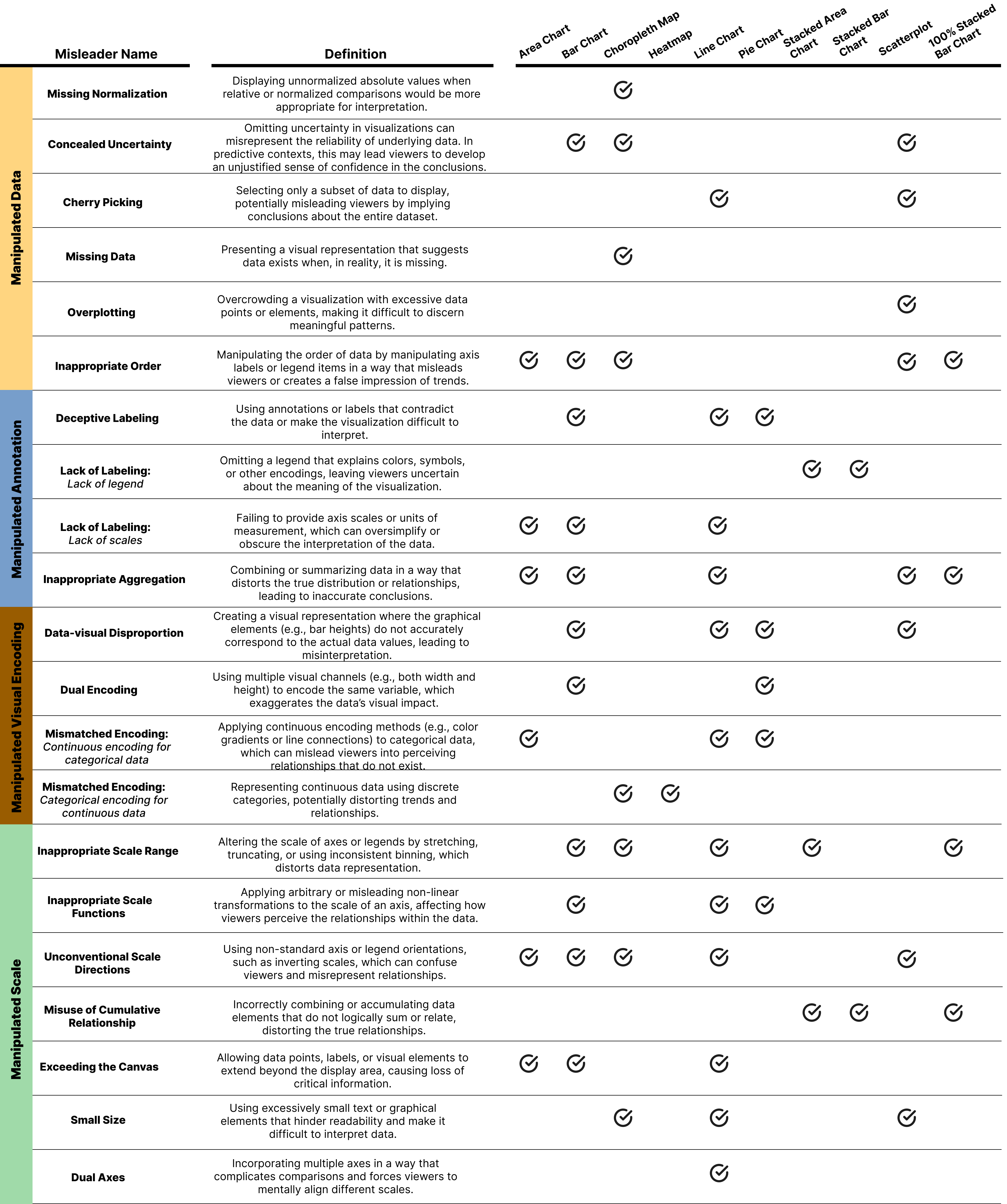}
\caption{List of misleaders categorized under each misleader group, along with their detailed definitions and corresponding chart types. In total, there are 60 (misleader, chart type) pairings.}
    \label{fig:misleaderDefinition}
\end{figure*}



\newpage
\subsection{Expert Labeling Tool Interface}

\begin{figure*}[!ht]
    \centering
    \includegraphics[width=\textwidth]{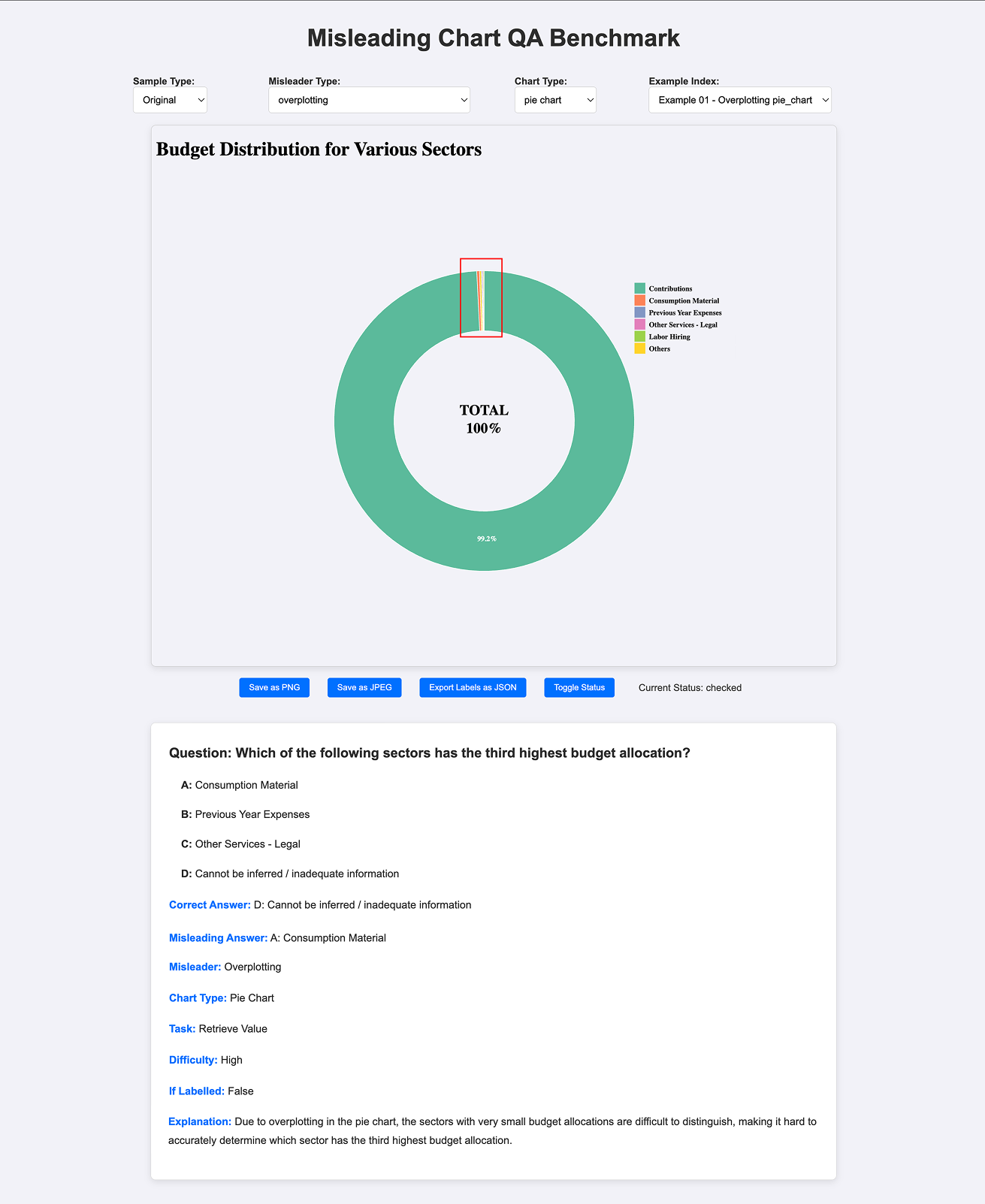}
\caption{Interface of our custom labeling tool used in the chart figure generation step. Experts annotate misleading regions using bounding boxes, as shown in the pie chart with an overplotting misleader. The interface also supports metadata editing, chart preview, and label export in standardized formats to facilitate expert validation and scalable dataset generation.}
    \label{fig:labellingTool}
\end{figure*}

\newpage

\twocolumn
\subsection{Expert Evaluation Guidelines}\label{guideline}
\subsubsection*{Overview}

To ensure high-quality outputs in the \textit{Misleading ChartQA} benchmark, each machine-generated MCQ was validated by PhD-level experts in data visualization. Experts used a custom labeling tool (Figure~\ref{fig:labellingTool}) to follow a structured 3-stage evaluation process guided by the protocol below.

\subsubsection*{Evaluation Protocol}

Please review each sample (including the chart, question, answer options, and explanation) following the steps below:

\begin{enumerate}
    \item \textbf{Verify Chart Correctness}
    \begin{itemize}
        \item Does the chart clearly and accurately demonstrate the intended misleader?
        \item Does it conform to the misleader definition in our taxonomy?
    \end{itemize}

    \item \textbf{Assess QA Pair Validity}
    \begin{itemize}
        \item Does the question clearly and accurately reflect the misleading aspect?
        \item Are the answer options logically sound?
        \item Does the marked correct answer resolve the question as intended?
        \item Does the marked misleading answer accurately reflect the misleading aspect as intended?
    \end{itemize}

    \item \textbf{Action Based on Assessment}
    \begin{itemize}
        \item \textit{\textcolor{red}{\textbf{Reject:}}} If the chart does not demonstrate the intended misleader, remove the sample.
        \item \textit{\textcolor{blue}{\textbf{Revise:}}} If the chart is correct but the QA pair is problematic (e.g., vague question, incorrect or ambiguous answers), revise the QA pair accordingly.
        \item \textit{\textcolor{green}{\textbf{Approve:}}} If both the chart and QA pair are accurate and coherent, approve without modification.
    \end{itemize}
\end{enumerate}

Each approved sample was confirmed by at least two independent experts, and revised samples underwent an additional round of expert validation.

\newpage
\subsection{Implementation Details of Experiments}\label{Implementation_Details}

Our experiments were conducted on 8 NVIDIA A$800$ GPUs (80GB each) using PyTorch $2$ and Python $3$. Given the task's complexity, we selected only the most advanced versions of each model type and evaluated them across different parameter sizes. Due to computational constraints, we randomly sampled around $30\%$ (876 cases) from the dataset for representativeness. 
\newpage

\onecolumn
\subsection{Comparison with related benchmarks}

\begin{table*}[!h]
\centering
    \resizebox{\textwidth}{!}{ 
\begin{tabular}{|cl|l|l|l|l|l|l|l|}
\hline
\multicolumn{2}{|c|}{\textbf{Task Focus}} & \multicolumn{1}{c|}{\textbf{Datasets}} & \multicolumn{1}{c|}{\textbf{\#-Chart Types}} & \multicolumn{1}{c|}{\textbf{\# Chart}} & \multicolumn{1}{c|}{\textbf{\# Task type}} & \multicolumn{1}{c|}{\textbf{Metadata?}} & \multicolumn{1}{c|}{\textbf{Chart Code?}} & \multicolumn{1}{c|}{\textbf{Chart Data?}} \\ \hline
\multicolumn{2}{|c|}{\multirow{2}{*}{Basic understanding}} & ChartQA & 3 & 4.8k & 4 & N & N & N \\ \cline{3-9} 
\multicolumn{2}{|c|}{} & PlotQA & 3 & 224k & 1 & N & N & N \\ \hline
\multicolumn{2}{|c|}{\multirow{4}{*}{Summarization/ captioning}} & ChartLlama & 10 & 11k & 7 & N & N & N \\ \cline{3-9} 
\multicolumn{2}{|c|}{} & ChartBench & 11 & 2.1k & 4 & N & N & N \\ \cline{3-9} 
\multicolumn{2}{|c|}{} & Chart-to-text & 6 & 44k & 3 & N & N & N \\ \cline{3-9} 
\multicolumn{2}{|c|}{} & Chartsumm & 3 & 84k & 1 & Y & N & N \\ \hline
\multicolumn{2}{|c|}{\multirow{2}{*}{Data/structure extraction}} & ChartInsights & 7 & 2k & 10 & Y & N & N \\ \cline{3-9} 
\multicolumn{2}{|c|}{} & FigureQA & 5 & 120K & 6 & N & N & N \\ \hline
\multicolumn{2}{|c|}{\textbf{Misleading Chart Comprehension}} & \textbf{Misleading ChartQA} & \textbf{10} & \textbf{3k} & \textbf{21} & \textbf{Y} & \textbf{Y} & \textbf{Y} \\ \hline
\end{tabular}
   }
\caption{Comparison of the Misleading ChartQA dataset with existing benchmarks. Misleading ChartQA is the first dataset specifically designed for the misleading chart comprehension task. It also features a diverse range of chart types and task types, along with rich metadata, chart code, and chart data.}
    \label{tab:comparison}
\end{table*}

\subsection{Chart Types Distribution}

\newpage

\subsection{Example: Manipulated Scale}\label{Manipulated_Scale}

\begin{figure*}[!h]
    \centering
    \includegraphics[width=0.75\linewidth]{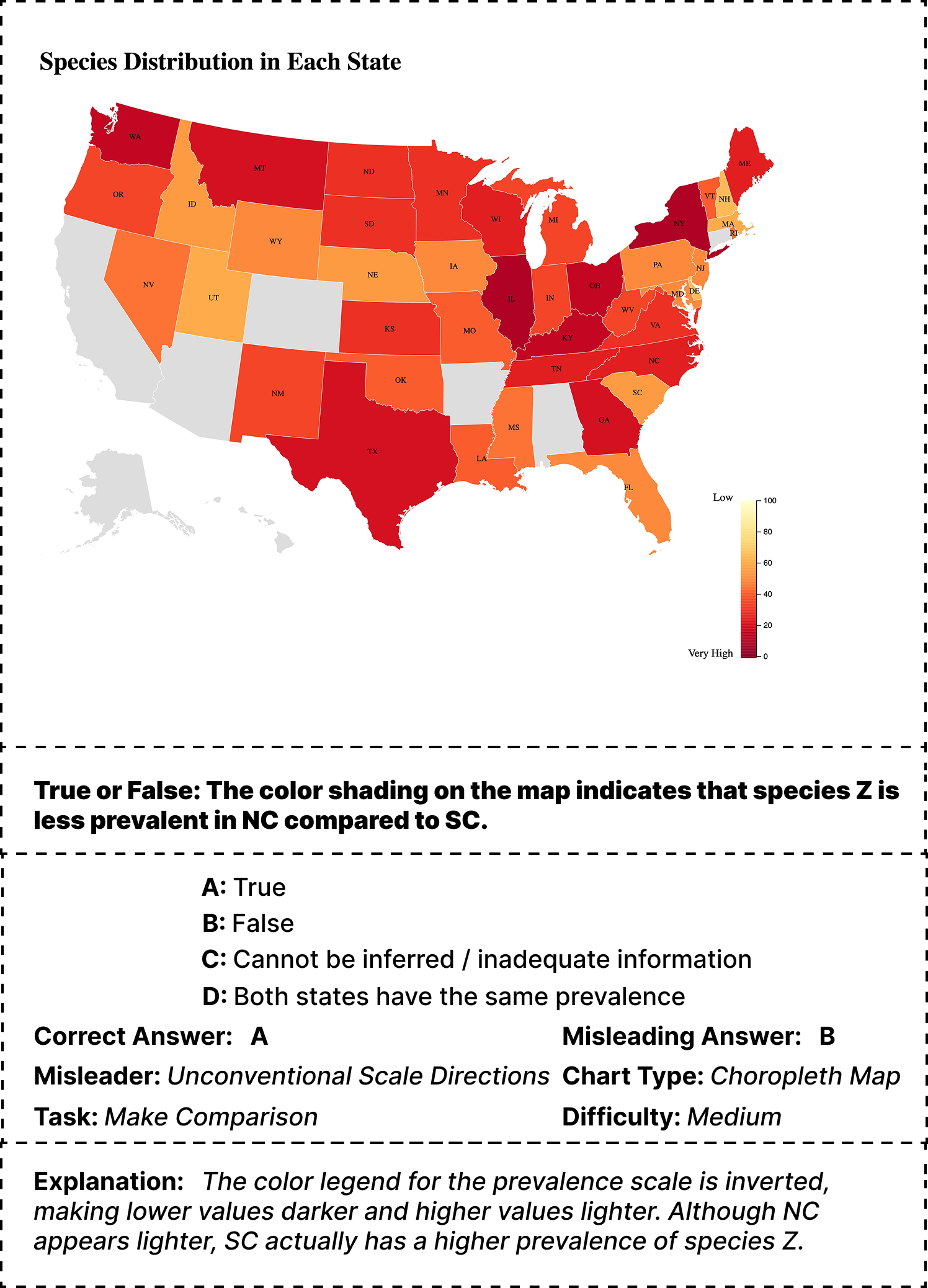}
        \caption{An example question from the \textbf{Manipulated Scale} group, categorized under \textit{Unconventional Scale Directions} and presented as a \textit{Choropleth Map}.}
    \label{fig:mani_scale_01}
\end{figure*}

\newpage
\begin{figure*}[!h]
    \centering
    \includegraphics[width=0.75\linewidth]{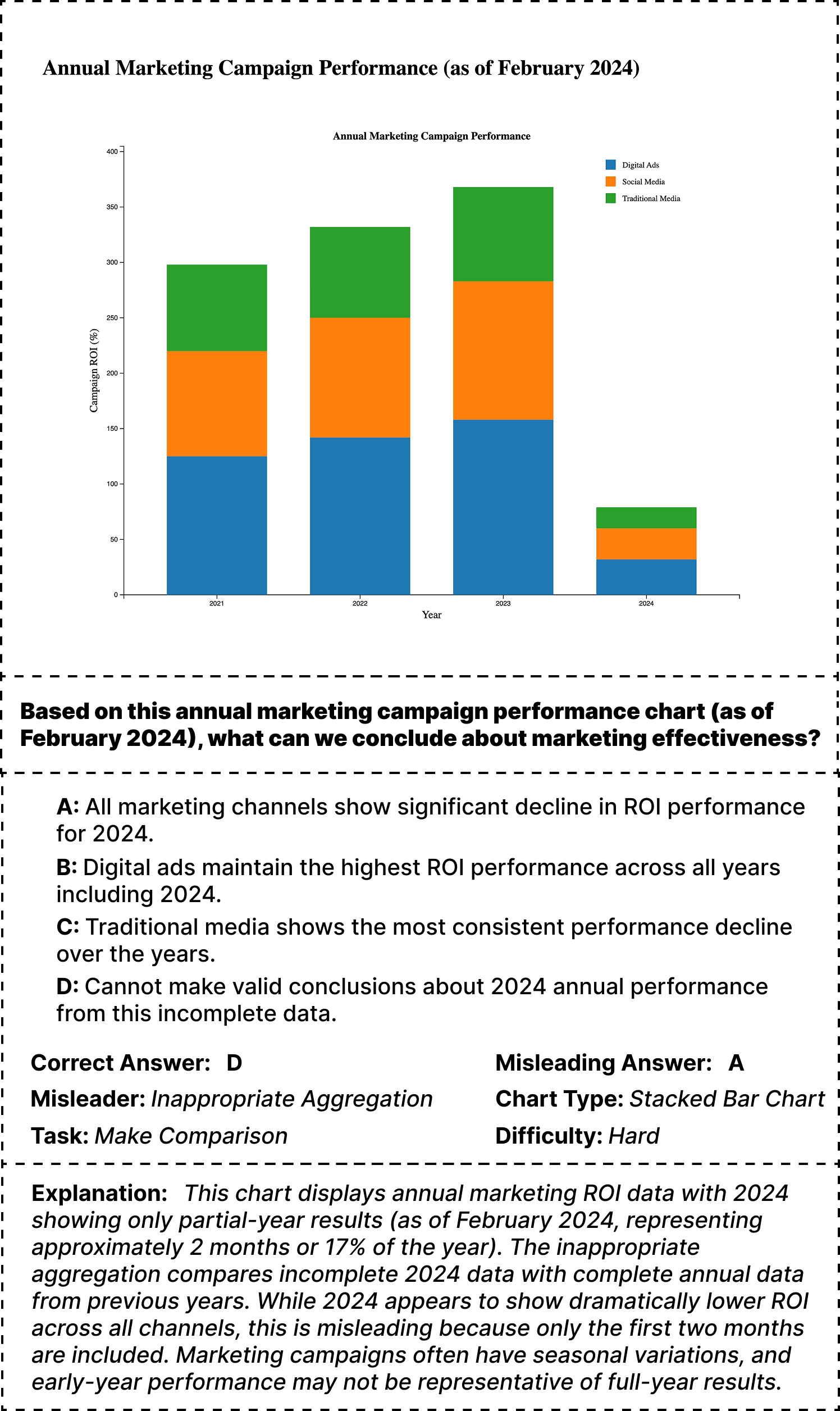}
    \caption{An example question from the \textbf{Manipulated Scale} group, categorized under \textit{Inappropriate Aggregation} and presented as a \textit{Stacked Bar Chart}.}
    \label{fig:mani_scale_02}
\end{figure*}

\newpage
\subsection{Example: Manipulated Annotation}\label{Manipulated_Annotation}

\begin{figure*}[!h]
    \centering
    \includegraphics[width=0.75\linewidth]{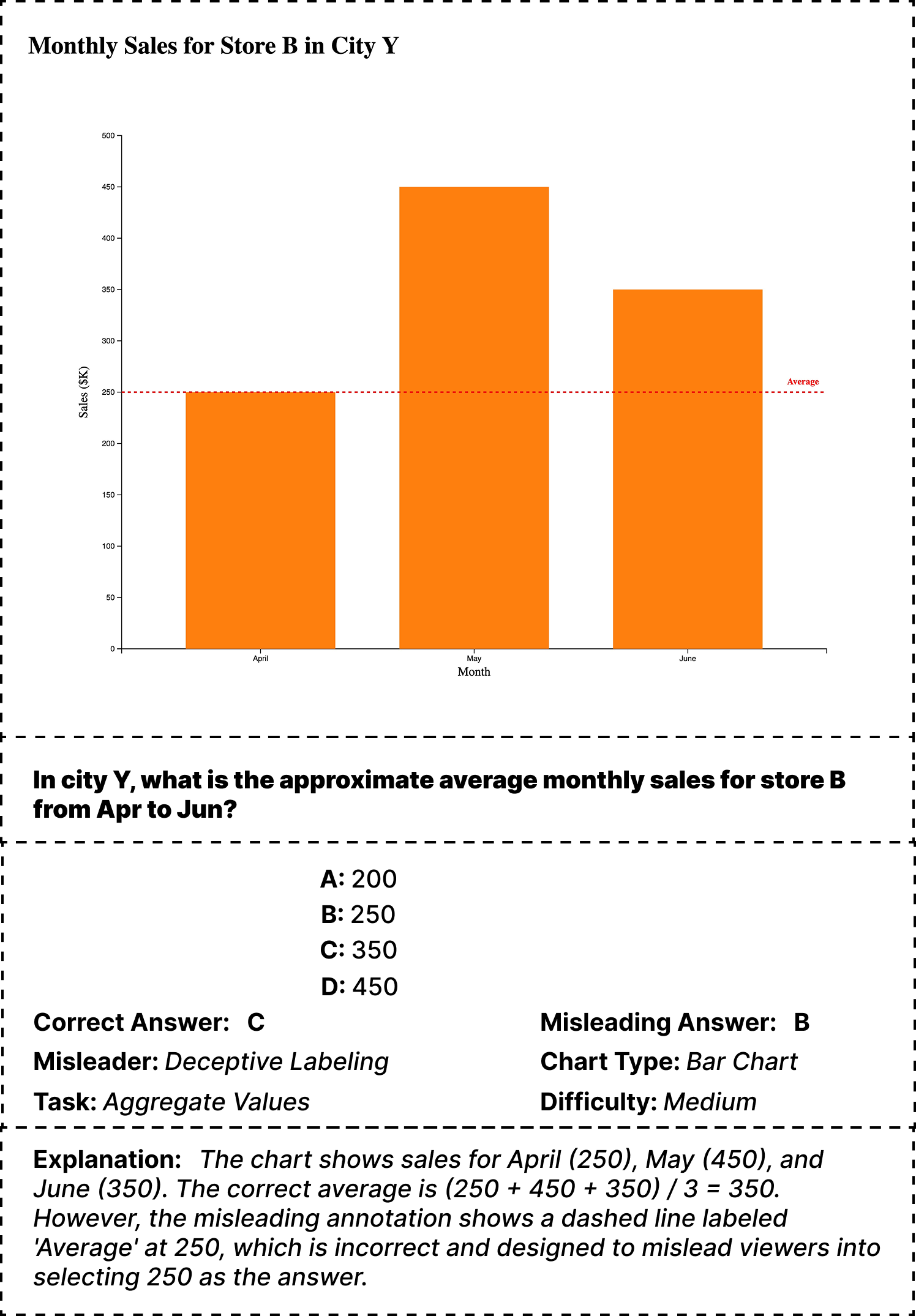}
        \caption{An example question from the \textbf{Manipulated Annotation} group, categorized under \textit{Deceptive Labelling} and presented as a \textit{Bar Chart}.}
    \label{fig:mani_anno_01}
\end{figure*}

\newpage
\begin{figure*}[!h]
    \centering
    \includegraphics[width=0.75\linewidth]{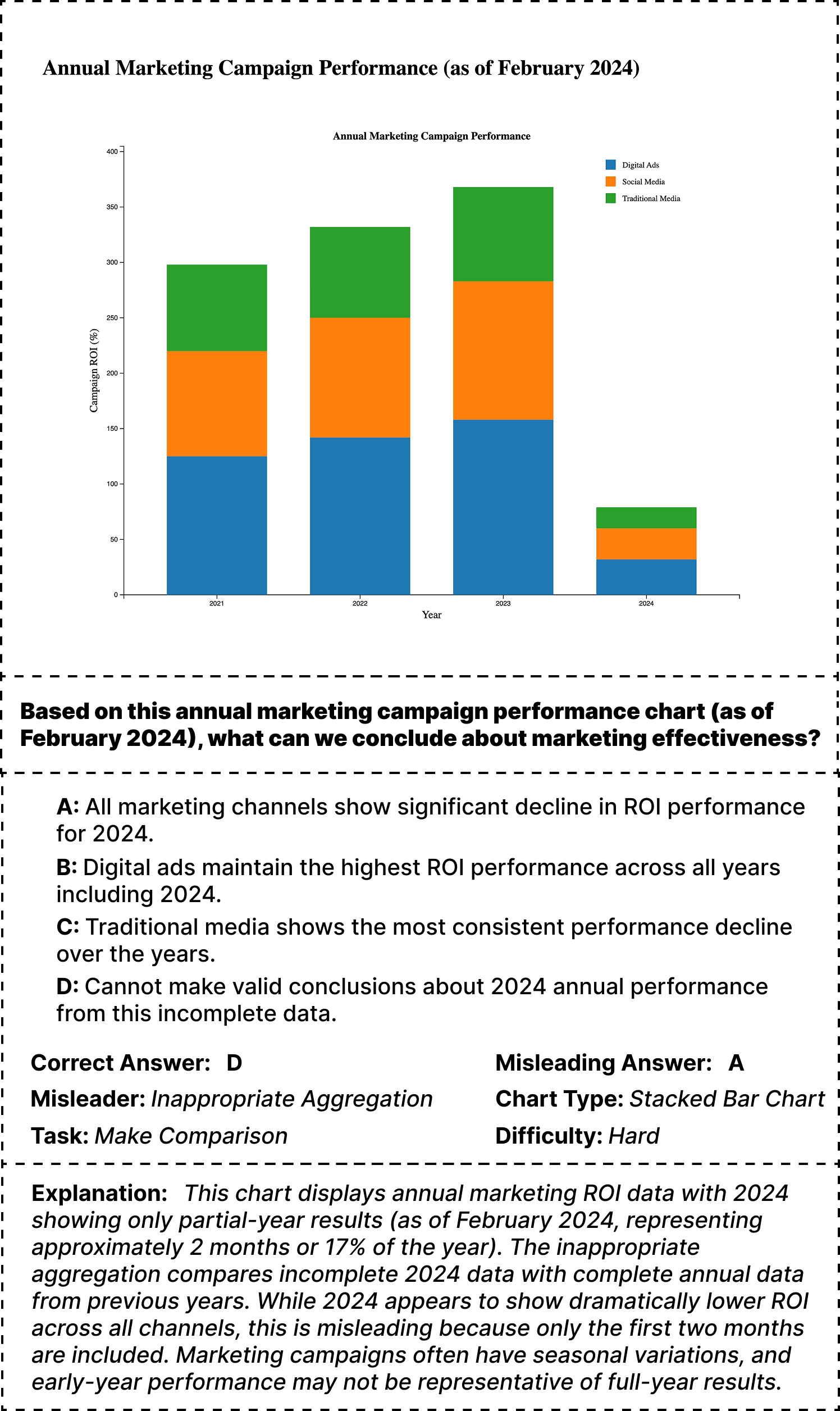}
    \caption{An example question from the \textbf{Manipulated Annotation} group, categorized under \textit{Inappropriate Aggregation} and presented as a \textit{Stacked Bar Chart}.}
    \label{fig:mani_anno_02}
\end{figure*}

\newpage
\subsection{Example: Manipulated Data}\label{Manipulated_Data}

\begin{figure*}[!h]
    \centering
    \includegraphics[width=0.75\linewidth]{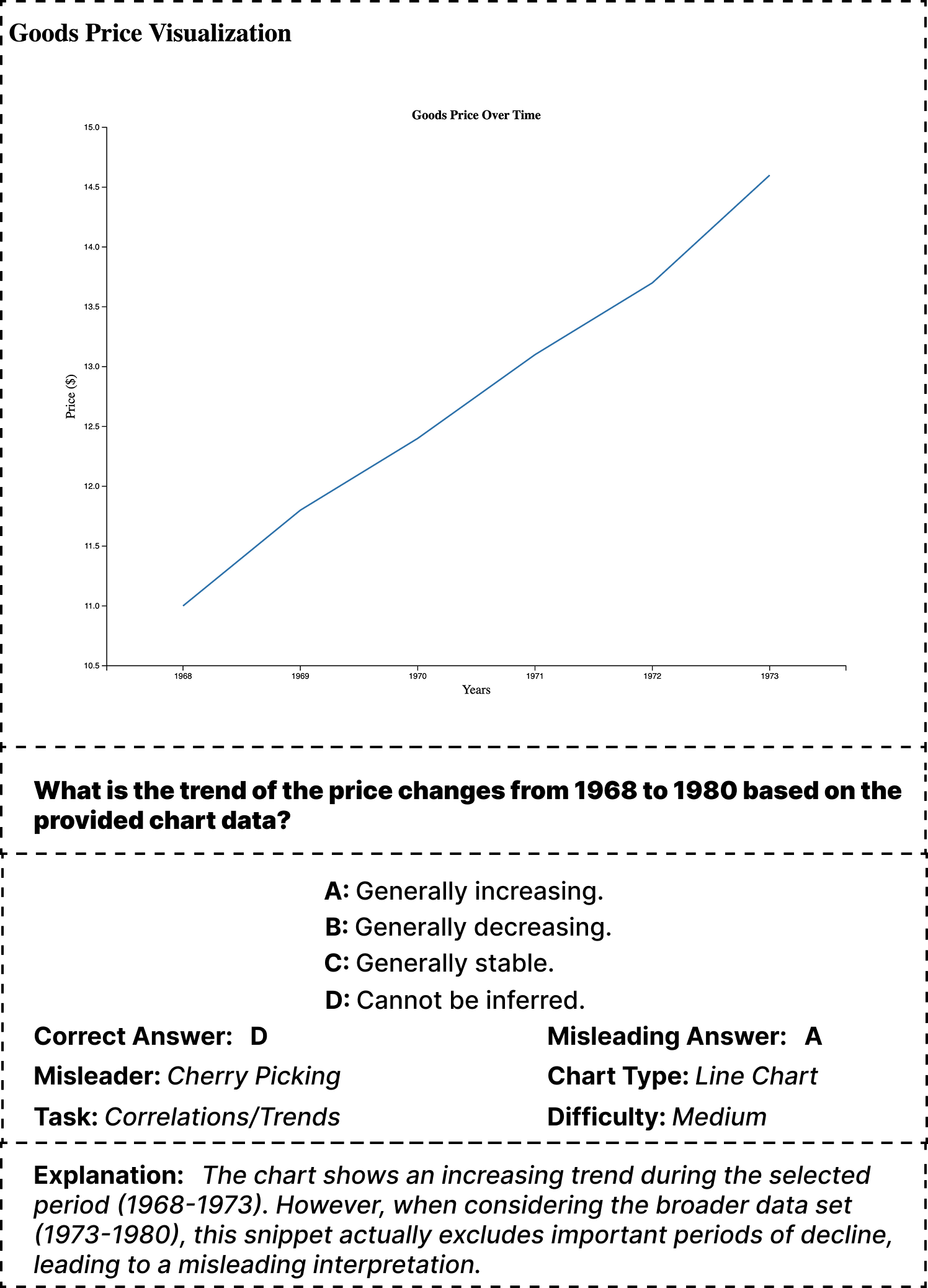}
    \caption{An example question from the \textbf{Manipulated Data} group, categorized under \textit{Cherry Picking} and presented as a \textit{Line Chart}.}
    \label{fig:cherrypick}
\end{figure*}
\newpage
\begin{figure*}[!h]
    \centering
    \includegraphics[width=0.75\linewidth]{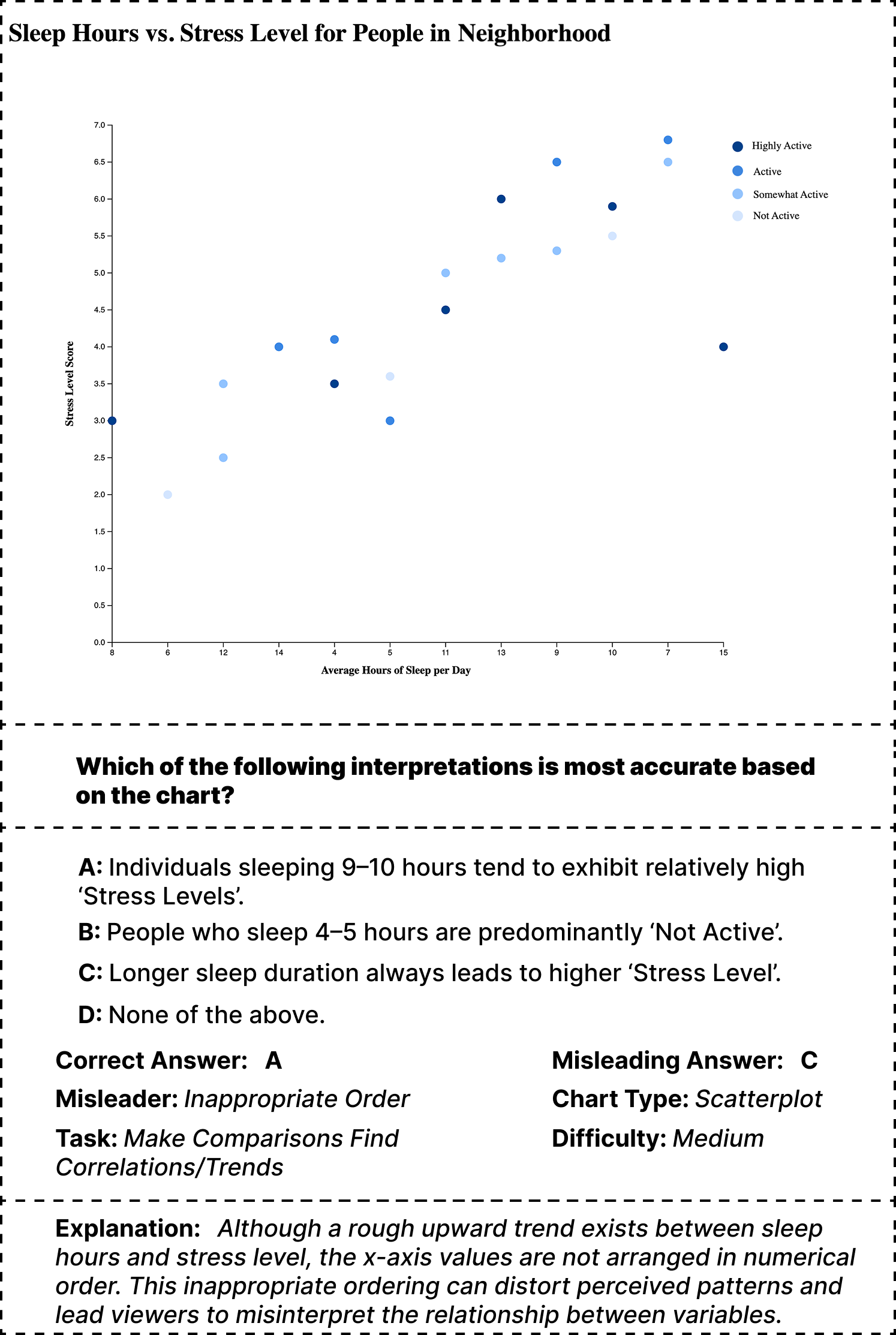}
        \caption{An example question from the \textbf{Manipulated Data} group, categorized under \textit{Inappropriate Order} and presented as a \textit{Scatterplot}.}
    \label{fig:inapporder}
\end{figure*}
\newpage

\subsection{Example: Manipulated Visual Encoding}\label{Manipulated_Visual_Encoding}

\begin{figure*}[!h]
    \centering
    \includegraphics[width=0.75\linewidth]{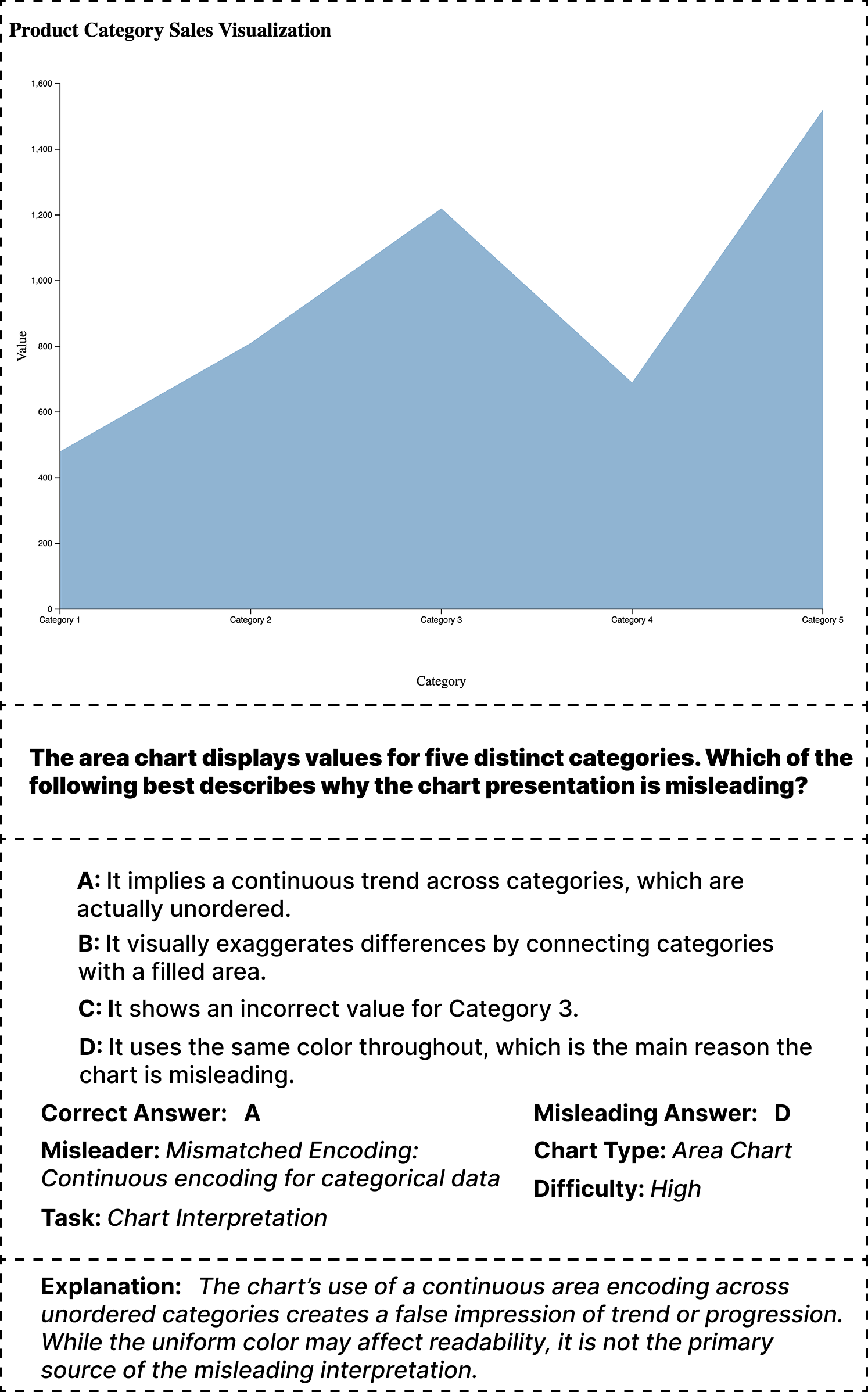}
        \caption{An example question from the \textbf{Manipulated Visual Encoding} group, categorized under \textit{Mismatched Encoding: Continuous encoding for categorical data} and presented as a \textit{Area Chart}.}
    \label{fig:mismatched}
\end{figure*}

\begin{figure*}[!h]
    \centering
    \includegraphics[width=0.75\linewidth]{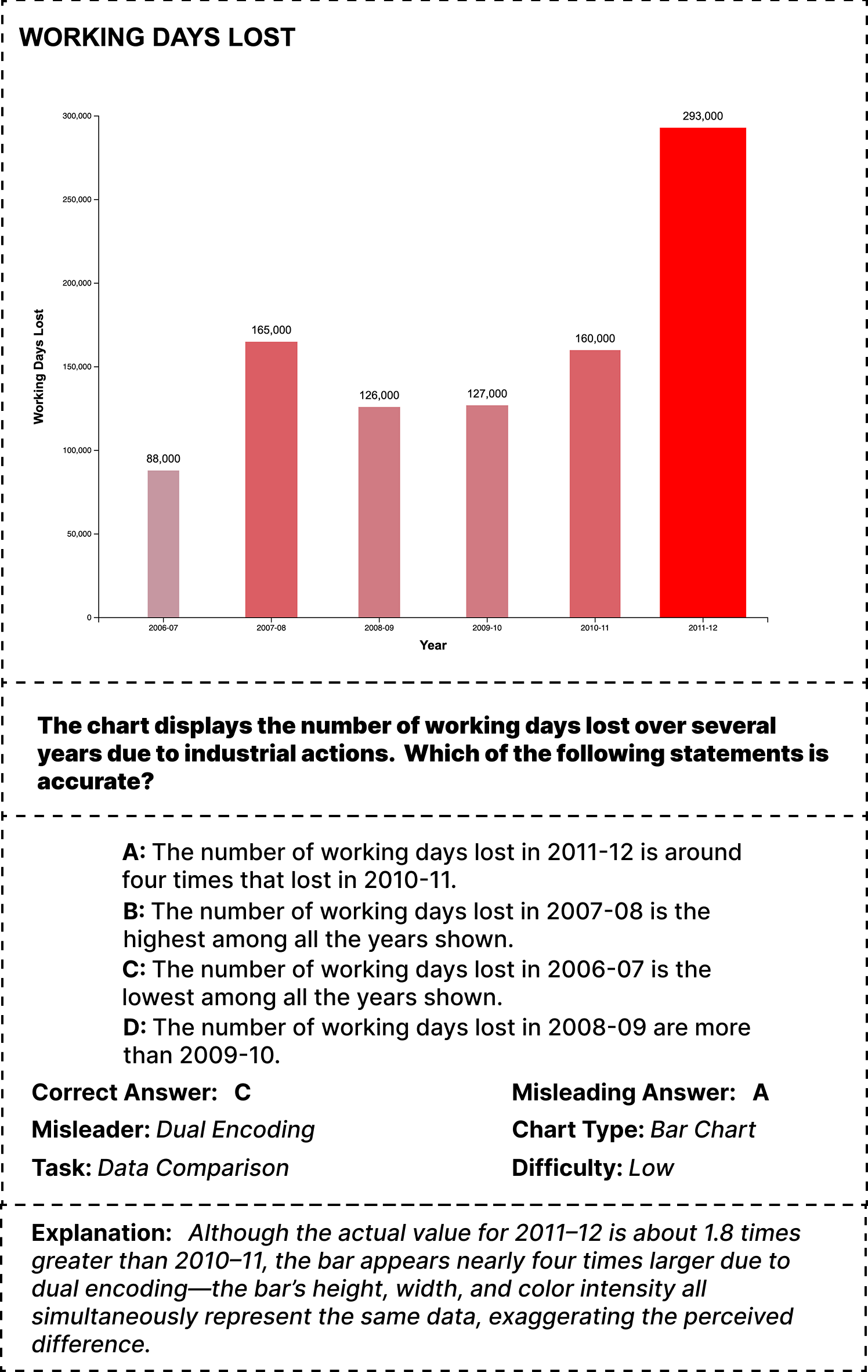}
    \caption{An example question from the \textbf{Manipulated Visual Encoding} group, categorized under \textit{Dual Encoding} and presented as a \textit{Bar Chart}.}
    \label{fig:datavisualdis}
\end{figure*}
\newpage

\newpage

\twocolumn
\subsection{Prompt Templates}\label{prompts}
\subsubsection{Automated MCQ Expansion and Iterative Refinement workflow}\label{prompt1}
The following are the prompts for each components in the proposed Automated MCQ Expansion and Iterative Refinement workflow (\cref{fig:generationPipeline}).

\begin{block}{Chart Variation}
\textit{\textbf{Generate HTML Variation}}\\
\noindent\rule{\linewidth}{0.4pt}

    You are generating misleading HTML-based charts for a QA benchmark using D3.js. The goal is to modify the visualization to reflect the misleader \highlight[orange]{$\{misleader\}$} by adjusting the chart's visual representation while maintaining core structure and labels. \\
    **Requirements:** \\
    1. The base HTML provided serves as the primary reference. Maintain the same overall structure, styles, and chart components. The generated HTML must be directly runnable. \\
    2. Retain the following from the base HTML: \\
       \noindent\hspace*{2em}- Chart dimensions (fixed at 1000x750 pixels). \\
       \noindent\hspace*{2em}- Titles, legends, axis labels, and grid lines. \\
       \noindent\hspace*{2em}- D3.js visualization logic. \\
    3. Modify the D3 chart to apply the misleader. \\
    4. Ensure the chart reads data from the updated CSV path: \highlight[orange]{$\{csv\_path\_in\_html\}$}. \\
       \noindent\hspace*{2em}- Ensure there are no extra or duplicated closing parentheses ')' in the 'd3.csv' function call. \\
    5. Prevent overflow by adjusting margins and ensuring all chart elements fit within the canvas. \\
    6. Use the labelled JPEG sample as a visual guide to ensure the misleader effect is accurately represented. \\
    7. Remove all unnecessary comments, such as: \\
       \noindent\hspace*{2em}- Descriptive comments like "Here's the complete and executable HTML page..." \\
       \noindent\hspace*{2em}- Markdown syntax (e.g., ```html, ```). \\
    8. **Ensure the chart title reflects the new chart topic but do not infer the misleader in the chart title**: \\
       \noindent\hspace*{2em}- The title should match the description of the relevant CSV columns. Make sure do not infer the misleaders in chart title. Keep the same  \\
    9. **Ensure axis labels dynamically update**: \\
       \noindent\hspace*{2em}- Use the column names from the CSV data for axis labels whenever appropriate. Make sure do not infer the misleaders in the axis labels. \\
    **Returns:** str: The generated HTML content only. \\
    \\
    **Misleader:** \highlight[orange]{$\{misleader\}$} \\
    **Misleader Description:** \highlight[orange]{$\{misleader\_description\}$} \\
    **Chart Type:** \highlight[orange]{$\{chart\_type\}$} \\
    **CSV Data (Driving the Chart):** \highlight[orange]{$\{csv\_data\}$} \\
    **Base HTML (Reference for Structure and Style):**
    \highlight[orange]{$\{base\_html\}$} \\
    **JPEG (Labelled Misleader):** \\
    \noindent\hspace*{2em}- Refer to the attached JPEG for visual alignment. 
    Path to JPEG: \highlight[orange]{$\{jpeg\_path\}$} \\
    **Ensure the full visualization code (chart headings, legends, titles, axes) is preserved:** \\
    **Return the output as a complete and executable HTML page** in the following format: \\
    \begin{verbatim}
     ``` 
    <!DOCTYPE HTML> 
    <html lang="en"> 
    <head> 
        <meta charset="UTF-8"> 
        <meta name="viewport" content="
        width=device-width, 
        initial-scale=1.0"> 
        <script src="https://d3js.org/
        d3.v6.min.js"></script>
        <style> 
            #chart {{ 
                width: 1000px; 
                height: 750px; 
                margin: 60px auto; 
            }} 
            .axis path, .axis line {{ 
                stroke: black; 
            }} 
            .dot {{
                fill: steelblue;
                stroke: black;
                stroke-width: 1px;
            }}
            .avg-line {{
                stroke: black;
                stroke-dasharray: 4,4;
            }}
            .annotation {{
                font-size: 12px;
                font-weight: bold;
                fill: black;
            }}
        </style>
    </head>
    <body>
        <h1> // // Insert appropriate chart 
        heading like the base HTML,
        ensure don't disclose the misleader 
        information here </h1>
        <div id="chart"></div>
        <script>
            // Insert D3.js visualization 
        logic extracted from base HTML here
        </script>
    </body>
    </html>
     ``` 
    \end{verbatim}
    \noindent\hspace*{2em}- Ensure that the returned HTML page preserves the full chart functionality and visualization logic from the base HTML.\\
    \noindent\hspace*{2em}- Implement the misleader described above by modifying axis scaling, bar order, or annotation placement.\\
    \noindent\hspace*{2em}- The goal is to introduce subtle distortions that create misleading visual interpretations while retaining the core chart layout.\\

\newpage

\textit{\textbf{Generate CSV Variation}}\\
\noindent\rule{\linewidth}{0.4pt}
    You are modifying CSV data for a \highlight[orange]{$\{chart\_type\}$} visualization that reflects the misleader \highlight[orange]{$\{misleader\}$}. \\
    **Instructions:**\\
    1. Keep the same number of columns (\highlight[orange]{$\{expected\_num\_columns\}$}) as the original CSV. \\
    2. Ensure each column has the same data type (e.g., int, float, string) as the original CSV.\\
    3. Modify column names and data values to reflect the misleader effect:\\
    \noindent\hspace*{2em}- \highlight[orange]{$\{misleader\_description\}$}\\
    4. Return only the modified CSV content with no additional comments or metadata. \\

    **Original CSV Data:**
    \highlight[orange]{$\{csv\}$}

\newpage

\end{block}

\begin{block}{QA Generation}

\textit{\textbf{Generate QA Variation}}\\
\noindent\rule{\linewidth}{0.4pt}
    You are generating Q\&A content for a misleading chart which is generated as a variation of the sample example. Please strictly follow the style of the sample (in which a chart with labeled misleading region and the corresponding Q\&A is provided). The goal is to craft a question that highlights the misleading aspect of the variation chart accordingly. \\

    **Requirements:**  \\
    1. Follow the structure of the provided JSON file exactly. \\
    2. Frame the question to reflect the misleading aspect of the chart. \\
    3. Adjust the options (A, B, C, D) to ensure one option aligns with the misleader. \\
    4. Indicate the correct answer clearly. \\
    5. Choose the most misleading option as "wrongDueToMisleaderAnswer" to highlight the most plausible incorrect option caused by the misleading chart. \\
    6. Reference the JPEG-labelled chart and Q\&A sample to ensure the explanation correctly addresses the visual misleader. \\
    7. Set the "ifLabelled" field to "False" to indicate the chart is not labelled. \\
    
    **Misleader:** \highlight[orange]{$\{misleader\}$} \\
    **Misleader Description:** \highlight[orange]{$\{misleader\_description\}$} \\
    **Chart Type:** \highlight[orange]{$\{chart\_type\}$} \\
    **CSV Data (Driving the Variation Chart):** \highlight[orange]{$\{csv\_data\}$} \\
    **The target Misleading Chart (Variation Chart):** \highlight[orange]{$\{chart\_variation\}$} \\
    **Sample Q\&A JSON (Structure Reference):**
    \highlight[orange]{$\{base\_json\}$} \\
    **Sample Chart JPEG (with Labelled Misleader):** \\
    \noindent\hspace*{2em}- Refer to the attached JPEG for visual alignment. \\
    \noindent\hspace*{2em}- Path to JPEG: \highlight[orange]{$\{jpeg\_path\}$} \\
    **Return the output in this strict format:**\\
    \begin{verbatim}
     ```json
    {{
      "question": "Based on the chart, 
      what is the approximate average sales 
      for Q1 2023 in Restaurant X?",
      "options": {{
        "A": "120",
        "B": "180",
        "C": "220",
        "D": "250"
      }},
      "correctAnswer": "B",
      "misleader": "{misleader}",
      "chartType": "{chart_type}",
      "task": "Aggregate Values",
      "explanation": "The chart annotation 
      shows 'Reference: 220', but the true 
      average is 180. Misleading 
      annotations cause users 
      to misjudge the data.",
      "difficulty": "Medium",
      "ifLabelled": "False",
      "wrongDueToMisleaderAnswer": "C"
    }}
    ```
    \end{verbatim}
    \end{block}
\newpage

\begin{block}{Automated Evaluation \& Feedback \& Refinement Loop}

\textit{\textbf{Variation Evaluation}}\\
\noindent\rule{\linewidth}{0.4pt}
     You are tasked with evaluating and refining a visualization QA sample for a misleading chart. \\

    ** Inputs  **\\
    \noindent\hspace*{2em}- **QA Content**: \highlight[orange]{$\{qa\_content\}$} \\
    \noindent\hspace*{2em}- **Misleader Description**: \highlight[orange]{$\{misleader\_desc\}$} \\
    \noindent\hspace*{2em}- **Misleadering Chart Image**: \highlight[orange]{$\{chart\_image\}$} \\
    \noindent\hspace*{2em}- **CSV Variation Check**: \highlight[orange]{$\{csv\_variation\_status\}$} \\
    \noindent\hspace*{2em}- **Generated CSV **: \highlight[orange]{$\{generated\_csv\}$} \\
    \noindent\hspace*{2em}- **Original CSV **: \highlight[orange]{$\{original\_csv\}$} \\
        
    ** Task **\\
        Evaluate the chart (visualization), question, QA options, correct answer, wrong-Due-To-Misleader-Answer all match the misleader description. If you find anything wrong, try to identify the corresponding errors in the CSV, QA, and HTML components based on the below guidelines and commen issues. \\
        Ensure: \\
        \noindent\hspace*{2em}- Make sure to double check the visualization indeed represents the intended misleader as described in the misleader description!\\ 
        \noindent\hspace*{2em}- Make sure to check if the QA content matches the misleader and visualization. \\
        \noindent\hspace*{2em}- Make sure to double check the correctness of the correct answer and wrongDueToMisleaderAnswer based on the misleader description and the chart figure! \\
        \noindent\hspace*{2em}- Make sure to check if the generated CSV introduces meaningful variations compared to the original CSV. \\
        \noindent\hspace*{2em}- Make sure to double check the items in the list of "Some common issues include" below. \\

        ** Guidelines **\\
        Evaluate the chart (visualization), question, QA options, correct answer, wrong-Due-To-Misleader-Answer, and alignment with the misleader description. Provide status as 'correct' or 'incorrect': \\
        \noindent\hspace*{2em}- "correct": No refinement needed. \\
        \noindent\hspace*{2em}- "incorrect": Refinement needed, provide comments and instructions. \\

        \noindent\hspace*{2em}- If the sample is correct, set "status": "correct" and leave "comments", "revision\_instructions", and "updated\_content" fields empty or as "No issues" and "null". \\
        \noindent\hspace*{2em}- If the sample requires refinement, set "status": "incorrect" and provide detailed comments and specific revision instructions for each component ("csv", "qa", "html"). \\

        ** For the updated\_content for "qa", directly provided the revised content in JSON format. **\\
        ** For the updated\_content for "csv" and "HTML", provide very detailed samples and do not include the whole code. **\\

        **  Some common issues include: **\\
            \noindent\hspace*{2em}**CSV:**\\
            \noindent\hspace*{2em}- The data values have no changes (no small variations) with the original data. Only changed the column names.\\
            \noindent\hspace*{2em}- Incorrect or missing data values.\\

           \noindent\hspace*{2em} **QA:**\\
            \noindent\hspace*{2em}- Mismatched question context (e.g., question does not align with the chart's content).\\
            \noindent\hspace*{2em}- Mismatched options (e.g., no correct answer choices exist).\\
            \noindent\hspace*{2em}- Missing or incorrect correct answers (e.g., no correct option, or wrong answer marked as correct).\\
            \noindent\hspace*{2em}- Incorrect explanations (e.g., explanation does not match the chart or the misleader description).\\
            \noindent\hspace*{2em}- Incorrect or missing wrongDueToMisleaderAnswer (e.g., wrong answer does not align with the misleader).\\
            
            \noindent\hspace*{2em}**HTML:**\\
            \noindent\hspace*{2em}- The CSV data path in the D3.js code is incorrect. Ensure the path in the D3.js code is path: \highlight[orange]{$\{csv\_path\_in\_html\}$}.\\
            \noindent\hspace*{2em}- Disclose the misleader in the visualization title (e.g., title implies it is a misleading visualization).\\
            \noindent\hspace*{2em}- Not specified by misleader description, but still missing labels or legend.\\
            \noindent\hspace*{2em}- Have any annotations to indicate misleading nature. Need to remove them.\\
        ** Output Format **\\
        Return a JSON object with the following structure:\\
    \begin{verbatim}
 ```json
    {{
        "status": "<correct/incorrect>",
        "comments": {{
            "csv": "<Comment for CSV 
            refinement or 'No issues'>",
            "qa": "<Comment for QA 
            refinement or 'No issues'>",
            "html": "<Comment for HTML 
            refinement or 'No issues'>"
        }},
        "revision_instructions": {{
            "csv": 
            "<Specific instructions 
            for revising the CSV or 
            'No revision required'>",
            "qa": 
            "<Specific instructions 
            for the revised QA or 
            'No revision required'>",
            "html": 
            "<Specific instructions 
            for revising the HTML or 
            'No revision required'>"
        }},
        "updated_content": {{
            "csv_data": "<Updated CSV 
            content if applicable or null>",
            "qa_content": "<Updated QA 
            content if applicable or null>",
            "html_content": "<Updated 
            HTML content if applicable 
            or null>"
        }}
    }}
```
    \end{verbatim}
\newpage

\textit{\textbf{Revision Loop: CSV}}\\
\noindent\rule{\linewidth}{0.4pt}
You are tasked with revising a CSV file to address specific issues. If you find no issues mentioned in the Comments and Instructions or they are unclear, please directlty output the Current CSV Content \highlight[orange]{$\{csv\_content\}$} without any changes. \\

*** Comments:\\
\highlight[orange]{$\{comments\}$}\\

*** Instructions:\\
\highlight[orange]{$\{instructions\}$}\\

*** Current CSV Content:\\
\highlight[orange]{$\{csv\_content\}$}\\

*** Revised CSV Sample:\\
\highlight[orange]{$\{revised\_csv\_sample\}$}\\

*** Task \\
Make the necessary revisions to the CSV file according to the Comments, Instructions and Revised CSV Sample. Return the updated content as a valid CSV file.\\

\newpage

\textit{\textbf{Revision Loop: HTML}}\\
\noindent\rule{\linewidth}{0.4pt}
You are tasked with revising an HTML file to address specific issues. If you find no issues mentioned in the Comments and Instructions or they are unclear, please directlty output the Current HTML Content \highlight[orange]{$\{html\_content\}$} without any changes. \\

*** Comments:\\
\highlight[orange]{$\{comments\}$}\\

*** Instructions:\\
\highlight[orange]{$\{instructions\}$}\\

*** Current HTML Content:\\
\highlight[orange]{$\{html\_content\}$}\\

*** Task \\
Make the necessary revisions to the HTML file and return the updated content as valid and executable HTML. \\
    \noindent\hspace*{2em}**Ensure the full visualization code (chart headings, legends, titles, axes) is preserved:**\\
    \noindent\hspace*{2em}**Make sure to replace the CSV path in the D3.js code with the correct path \highlight[orange]{$\{csv\_path\_in\_html\}$}.**\\
    \noindent\hspace*{2em}**Make sure to remove any annotations or titles in the visualization that disclose the misleader! (e.g., should not have some extra titles indicating the potential misleader)**\\
    \noindent\hspace*{2em}**Make sure the visualization represents the misleader as intended.**\\
    \noindent\hspace*{2em}**Make sure to not change the other parts of the visualization code.**\\
    \noindent\hspace*{2em}**Return the output as a complete and executable HTML page** in the following format: \\

    \begin{verbatim}
    ```
    <!DOCTYPE html>
    <html lang="en">
    <head>
        <meta charset="UTF-8">
        <meta name="viewport" content=
        "width=device-width, 
        initial-scale=1.0">
        <script src="https://d3js.org/
        d3.v6.min.js"></script>
        <style>
            #chart {{
                width: 1000px;
                height: 750px;
                margin: 60px auto;
            }}
            .axis path, .axis line {{
                stroke: black;
            }}
            .dot {{
                fill: steelblue;
                stroke: black;
                stroke-width: 1px;
            }}
            .avg-line {{
                stroke: black;
                stroke-dasharray: 4,4;
            }}
            .annotation {{
                font-size: 12px;
                font-weight: bold;
                fill: black;
            }}
        </style>
    </head>
    <body>
        <h1> // Insert appropriate chart 
        heading like the base HTML, 
        ensure do not 
        disclose the misleader 
        information here </h1>
        <div id="chart"></div>
        <script>
            // D3.js visualization logic
            d3.csv("{csv_path_in_html}")
                .then(function(data) {{
                    // Chart logic here
                }})
                .catch(function(error) {{
                    console.error('Error 
                    loading CSV data:', 
                    error);
                }});
        </script>
    </body>
    </html>
    ```
    \end{verbatim}

\newpage
\textit{\textbf{Revision Loop: Q\&A}}\\
\noindent\rule{\linewidth}{0.4pt}
You are tasked with revising a QA JSON file to address specific issues. If you find no issues mentioned in the Comments and Instructions or they are unclear, please directlty output the Current QA Content \highlight[orange]{$\{qa\_content\}$} without any changes. \\

*** Comments:\\
\highlight[orange]{$\{comments\}$}\\

*** Instructions:\\
\highlight[orange]{$\{instructions\}$}\\

*** Current QA Content:\\
\highlight[orange]{$\{qa\_content\}$}\\

***  Revised QA Recommendation:\\
\highlight[orange]{$\{revised\_qa\_recommendation\}$}\\

*** Task \\
Make the necessary revisions to the QA JSON file and return the updated content as valid JSON.

**Return the output in this strict format:** \\
    \begin{verbatim}
```json
{{
    "question": "Based on the chart, what 
    is the approximate average sales for 
    Q1 2023 in Restaurant X?",
    "options": {{
    "A": "120",
    "B": "180",
    "C": "220",
    "D": "250"
    }},
    "correctAnswer": "B",
    "misleader": "misleader",
    "chartType": "chart_type",
    "task": "Aggregate Values",
    "explanation": "The chart annotation 
    shows 'Reference: 220', but the true 
 average is 180. Misleading annotations 
    cause users to misjudge the data.",
    "difficulty": "Medium",
    "ifLabelled": "False",
    "wrongDueToMisleaderAnswer": "C" }}
``` \end{verbatim}

\end{block}

\newpage
\subsubsection{Prompt Templates for the Main Experiments}\label{prompt2}
The following are the prompt templates for the \textbf{Baseline} and \textbf{Zero-shot CoT} experimental settings (\cref{tab:summary_result}).
\begin{block}{Baseline}
\textit{\textbf{Core Prompts for Baseline Experiment}}\\
\noindent\rule{\linewidth}{0.4pt}
    You are given a potentially misleading chart and a multiple-choice question related to it. Please provide the MCQ answer and the corresponding explanation: \\
        
    ** The Potentially Misleading  Chart: ** \highlight[orange]{$\{image\_path\}$} \\
    ** Question: **
        \highlight[orange]{$\{question\}$} \\
    ** Options: ** 
        \highlight[orange]{$\{formatted\_options\}$} \\

    ** Instructions: ** \\
        \noindent\hspace*{2em}- **Only output the selected option on the first line (A, B, C, or D).** \\
        \noindent\hspace*{2em}- Then, on a new line, **provide a detailed explanation** on why this choice is correct based on the chart. \\
        
        \noindent\hspace*{2em}- Your response format must strictly follow: \\
        \noindent\hspace*{4em}<Letter Choice> \\
        \noindent\hspace*{4em}<Explanation> \\
        \noindent\hspace*{2em}- For example: \\
    \begin{verbatim}
        ```
        B
        The price trend is decreasing from 
        1975 to 1980, as the line clearly 
        slopes downward.
        ```
    \end{verbatim}
    Now, answer accordingly, do not forget to provide the explanation for your answer: \\
    
    \newpage

\end{block}

\begin{block}{Zero-shot CoT}
\textit{\textbf{Core Prompts for Zero-shot CoT Experiment}}\\
\noindent\rule{\linewidth}{0.4pt}
    You are given a potentially misleading chart and a multiple-choice question related to it. Please provide the MCQ answer and the corresponding explanation. ** Let's think and solve the question step by step!** \\

    ** The Potentially Misleading Chart: ** \highlight[orange]{$\{image\_path\}$} \\
    ** Question: **
        \highlight[orange]{$\{question\}$} \\
    ** Options: ** 
        \highlight[orange]{$\{formatted\_options\}$} \\

    ** Instructions: ** \\
        \noindent\hspace*{2em}- **Start with breaking down the problem and think through the question logically. \\
        \noindent\hspace*{2em}- **You can first try to analyze the chart components (e.g., chart title, chart axis, ...), then based on the chart analysis, continue with the analysis of QA. \\
        \noindent\hspace*{2em}- After reasoning, output the selected option (A/B/C/D) and explain your choice based on the chart.\\

    ** Please Ensure: ** \\
        \noindent\hspace*{2em}- **Only output the selected option on the first line (A, B, C, or D).** \\
        \noindent\hspace*{2em}- Then, on a new line, **provide a detailed explanation** on why this choice is correct based on the chart. \\        
        \noindent\hspace*{2em}- Your response format must strictly follow: \\
        \noindent\hspace*{4em}<Letter Choice> \\
        \noindent\hspace*{4em}<Explanation> \\
        \noindent\hspace*{2em}- For example: \\
    \begin{verbatim}
        ```
        B
        The price trend is decreasing from 
        1975 to 1980, as the line clearly 
        slopes downward.
        ```
    \end{verbatim}
    Now, answer accordingly, do not forget to provide the explanation for your answer: \\
    
    \newpage

\end{block}

\newpage
\subsubsection{Region-Aware Misleading Chart Reasoning Pipeline}\label{prompt3}
The following are the prompts for each components in the proposed Region-Aware Misleading Chart Reasoning pipeline (\cref{fig:pipeline}).
\begin{block}{Misleading Region Identification}
\textit{\textbf{MLLM Module for Misleading Region Identification}}\\
\noindent\rule{\linewidth}{0.4pt}
       You are given a chart (dimensions: 2400 x 2122) with potential misleading regions: \highlight[orange]{$\{image\_path\}$} \\
 
        Please analyze the image to detect any misleading regions (e.g., the chart design or data select might be intentionally manipulate the data's visual representation to bolster specific claims, 
        can distort viewers' perceptions and lead to decisions rooted in false information).  \\

        ** Let's think it step by step! ** Here is a potential checklist for identifying misleading regions that you may refer to: \\\\
        \noindent\hspace*{2em}- Chart Title \\
        \noindent\hspace*{2em}- Chart Type \\
        \noindent\hspace*{2em}- X and Y Axis\\
        \noindent\hspace*{2em}- Chart Legend\\
        \noindent\hspace*{2em}- Chart Visual Encoding\\
        \noindent\hspace*{2em}- Chart Data Use and Choice\\
        \noindent\hspace*{2em}- Chart Scales\\
        \noindent\hspace*{2em}- Chart Annotations\\
        
        Then output a JSON file containing coordinates for the potential misleaders and explanations. \\

        *** Instructions:
        \noindent\hspace*{2em}- **Please analyze the image (dimensions: 2400 x 2100) to detect any misleading regions.** \\
        \noindent\hspace*{2em}- **Provide the misleading region coordinates with a detailed explanation** \\
        \noindent\hspace*{2em}- Your response format must strictly follow the example JSON format: \\

    \begin{verbatim}
        ```
        [
            {{"coordinates": [[100, 200], 
            [150, 200],[100, 300], 
            [150, 300]], 
            "explanation": "The chart 
            incorrectly scales 
            the y-axis."}},
            {{"coordinates": [[250, 300], 
            [300, 300],[250, 350], 
            [300, 350]], 
            "explanation": "The chart uses 
            misleading colors that 
            misrepresent data."}}
        ]
        ```
    \end{verbatim}

    \newpage

\end{block}

\begin{block}{Q\&A with Labeled Reference Region}
\textit{\textbf{MLLM Module for Q\&A with Labeled Reference Region}}\\
\noindent\rule{\linewidth}{0.4pt}
    You are given a chart with potential misleading regions and a corresponding question. Additionally, you will receive an extra image where the potential misleading region is labeled with an explanation. Use this as a reference, ** but please note that the labels may not always be accurate! ** Answer the question with a clear explanation. \\

    ** The original Chart: ** \highlight[orange]{$\{image\_path\}$} \\
    ** Question: **
        \highlight[orange]{$\{question\}$} \\
    ** Options: ** 
        \highlight[orange]{$\{formatted\_options\}$} \\
    ** The labeled Chart: ** \highlight[orange]{$\{labeled_image\_path\}$} \\    
     ** Explanations for the labels: ** 
        \highlight[orange]{$\{regions\_explanation\}$}
 
    ** Instructions: ** \\
        \noindent\hspace*{2em}- **Only output the selected option on the first line (A, B, C, or D).** \\
        \noindent\hspace*{2em}- Then, on a new line, **provide a detailed explanation** on why this choice is correct based on the chart. \\

        \noindent\hspace*{2em}- Your response format must strictly follow: \\
        \noindent\hspace*{4em}<Letter Choice> \\
        \noindent\hspace*{4em}<Explanation> \\
        \noindent\hspace*{2em}- For example: \\
    \begin{verbatim}
        ```
        B
        The price trend is decreasing from 
        1975 to 1980, as the line clearly 
        slopes downward.
        ```
    \end{verbatim}
    Now, answer accordingly: \\
    
    \newpage

\end{block}



\end{document}